\documentclass[sigconf]{acmart}
\AtBeginDocument{%
  \providecommand\BibTeX{{%
    \normalfont B\kern-0.5em{\scshape i\kern-0.25em b}\kern-0.8em\TeX}}}

\usepackage[utf8]{inputenc} 
\usepackage[T1]{fontenc}    
\usepackage{hyperref}       
\usepackage{url}            
\usepackage{booktabs}       
\usepackage{amsfonts}       
\usepackage{nicefrac}       
\usepackage{microtype}      
\usepackage{array}
\usepackage{graphics}
\usepackage{graphicx}
\usepackage{algorithm,algorithmicx,amsmath}
\usepackage[noend]{algpseudocode}
\usepackage{appendix}
\usepackage{balance}

\begin{document}
\setlength{\headheight}{16pt}
\settopmatter{printacmref=false}
\setcopyright{none}
\renewcommand\footnotetextcopyrightpermission[1]{}
\pagestyle{plain}
\title{Personalized Federated Deep Learning\texorpdfstring{\\}{} for Pain Estimation From Face Images}

\author{Ognjen (Oggi) Rudovic*\texorpdfstring{$^{1,2}$}{}, Nicolas Tobis*\texorpdfstring{$^{2}$}{}, Sebastian Kaltwang\texorpdfstring{$^{2}$}{}, Bj{\"o}rn Schuller\texorpdfstring{$^{2}$}{}, Daniel Rueckert\texorpdfstring{$^{2}$}{}, Jeffrey F. Cohn\texorpdfstring{$^{3}$}{} and Rosalind W. Picard\texorpdfstring{$^{1}$}{}}
\thanks{O.Rudovic and N.Tobis contributed equally to this work.}
\affiliation{{\small  $^1$Massachusetts Institute of Technology, MA, USA, $^2$Imperial College London, UK, $^3$Department of Psychology, PA, USA}\\{\tt\small \{orudovic, roz\}@mit.edu,\{nicolas.tobis18, sebastian.kaltwang08, bjoern.schuller, d.rueckert\}@imperial.ac.uk, jeffcohn@pitt.edu}}

\renewcommand{\shortauthors}{ }

\begin{abstract}
  Standard machine learning approaches require centralizing the users' data in one computer or a shared database, which raises data privacy and confidentiality concerns. Therefore, limiting central access is important, especially in healthcare settings, where data regulations are strict.
  A potential approach to tackling this is Federated Learning (FL), which enables multiple parties to collaboratively learn a shared prediction model by using parameters of locally trained models, while keeping raw training data {\it locally} (e.g., on a mobile device). In the context of AI-assisted pain-monitoring, we wish to enable confidentiality-preserving and unobtrusive pain estimation for long-term pain-monitoring, and reduce the burden on the nursing staff who perform frequent routine check-ups. To this end, we propose a novel Personalized Federated Deep Learning (PFDL) approach for pain estimation from face images. PFDL performs collaborative training of a deep model, implemented using a lightweight CNN architecture, across different clients (i.e., subjects) without sharing their face images. Instead of sharing all parameters of the model, as in standard FL, PFDL retains the last layer locally (used to personalize the pain estimates). This (i) adds another layer of data confidentiality, making it difficult
  for an adversary to infer pain levels of the target subject, while (ii) enabling to personalize the pain estimation to each subject through local parameter tuning. We show using a publicly available dataset of face videos of pain (UNBC-McMaster Shoulder Pain Database), that PFDL performs comparably or better than the standard centralized and FL algorithms, while further enhancing data privacy. This, in turn, has potential to improve the traditional pain monitoring by making it more secure, computationally efficient and scalable to a large number of individuals (e.g., for in-home pain monitoring), for whom we aim to provide timely and unobtrusive pain measurement.
\end{abstract}

\settopmatter{printfolios=true}
\maketitle

\begin{figure*}
    \centering
    \includegraphics[width=.7\textwidth]{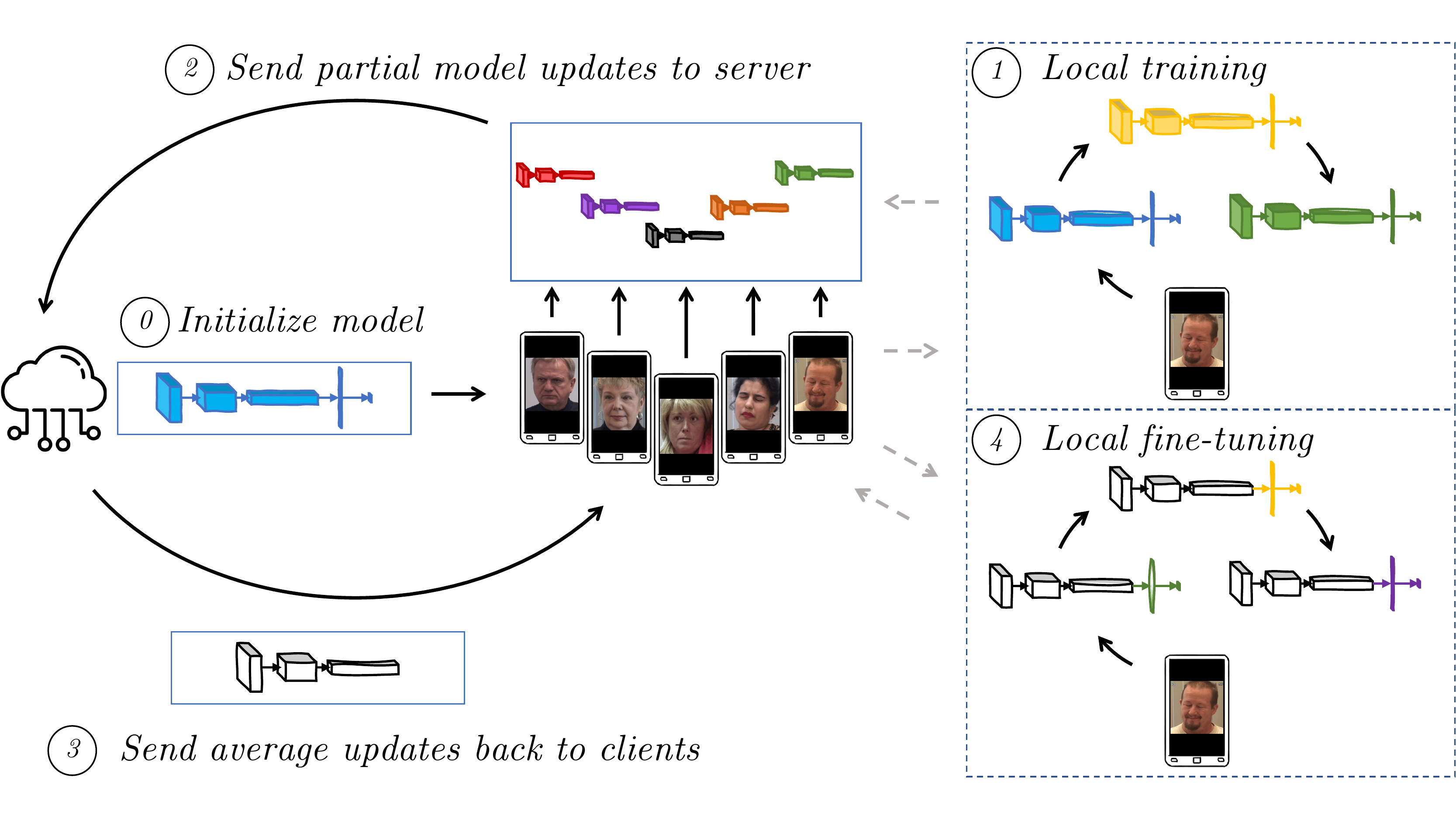}
    \caption{Personalized Federated Deep Learning for Pain Estimation From Face Images. Initially, the deep model is pre-trained on available data of non-target subjects (Step 0). The parameters of this model are then used to initialize the target-subject-specific models, which are trained using the subject data locally (Step 1). The parameters of the local (individual) models -- the convolutional layers but not the last two fully-connected dense layers (thus further securing the confidentiality of the subjects) -- are updated using the federated averaging approach (Step 2). These updates are then returned to local sites (Step 3) and only the two dense layers are further fine-tuned for estimation of the subject-specific pain expressions (Step 4). The steps 1-4 are repeated on-the-fly for the new recording sessions, as depicted in Fig.~\ref{fig:sessions}.}
    \label{fig:federated-personalization}
\end{figure*}

\section{Introduction}
While the potential applications of Federated Learning (FL)~\cite{ref-google-FML-init} are numerous and highly diverse, the promise of a confidentiality-preserving\footnote{While {\it standard} Federated Learning does not provide privacy guarantees (in the sense that it is not possible for an adversary to infer properties of the data that is considered private); it provides confidentiality guarantees (i.e., the data is not accessible to users that are not authorized to access it). In the context of this work we refrain from using the term privacy-preserving, and use the term confidentiality-preserving instead. However, to our knowledge there is no prior work that succeeded in reconstructing user's face images from a deep-model parameters only, and we argue that it would be extremely difficult to achieve this using the subset of the parameters shared by our PFDL model (see Sec.\ref{sec:fl} for more details).}, less computationally demanding machine learning approach, can be particularly valuable for the healthcare space. Healthcare data are typically among the most strictly regulated with HIPAA~\cite{hipaa} in the US and GDPR~\cite{gdpr} in the European Union, governing the strict rules by which such data can be accessed/shared. At the same time, pooling personal health data from healthcare facilities, insurance providers, and government agencies on a regional, national or even global scale holds enormous potential, for example, for research on rare diseases~\cite{rare-disease}, through clinical trials or treatment best-practices, such as automated pain monitoring~\cite{hammal2012pain}. The latter, for instance, can inform the nursing stuff in a timely way about a patients' response to medication, and, thus, allow them to adjust the treatment through personalized delivery of the target medication~\cite{cervero2012understanding}. This could potentially improve the quality of care and life for patients, and help reduce risks related to both pain-reliever underdosing and overdosing.

Research on automated pain monitoring has been facilitated by the availability of publicly available datasets~\cite{werner2019automatic}, especially of facial expressions of pain (the UNBC-McMaster dataset~\cite{painful-data}). Over the last decade, various machine learning (ML) and computer vision (CV) models have been proposed for pain estimation using this face-videos dataset (e.g., ~\cite{ashraf2009painful, lucey2010automatically, kaltwang2012continuous,facial-action-unit}). So far, these methods have explored only the centralized learning of models for pain estimation, which inevitably requires sharing raw personal data (face images). Data sharing across multiple sites (hospitals or mobile devices) can greatly increase the amount of data for training ML models, which helps improve the performance of the pain estimation systems. Nevertheless, this raises questions about data confidentiality, accountability and governance, especially if such systems are to be implemented in real-world settings. To address this, we use FL~\cite{ref-google-FML-init} to design a novel framework for distributed deep learning of pain estimation models from face images. Specifically, we propose a new learning algorithm that preserves the user's confidentiality by never sharing their raw personal data, but only their model parameters during the model training. To further prevent a potential misuse of private data\footnote{In this work, with "private data" we refer to the user's face images and corresponding pain labels.}, the model parameters required for estimation of the user's pain (the classification layer in our model) are also never shared; instead, they are optimized locally using a small portion of the target user's pain data. 
The contributions of this work can be summarized as follows:
\begin{itemize}
    \item To the best of our knowledge, this is the first work that addresses the {\it confidentiality-preserving} estimation of pain from face images. Using a challenging dataset of real-world spontaneously displayed facial expressions of pain ("UNBC-McMaster Shoulder Pain Expression Archive"), we show that the performance of personalized FL with deep models for pain estimation is similar or better (in terms of the F1 score for classification of pain/no pain in test images) than the traditional centralized learning. The key advantage of our approach is that it preserves data confidentiality - the user's data (face images and pain labels) are always retained locally (e.g., on the user's camera-enabled device).   
    \item We introduce a novel modification of the traditional FL algorithm, termed {\it federated personalization}, that adds an additional level of data protection. Instead of using all the model parameters for communication rounds between the user-specific models, our approach retains the classification layer in the deep model for pain estimation. This makes it hard for an adversary to infer the user's expressions of pain, while allowing the model to {\it personalize} its estimations to the target user.
\end{itemize}
The outline of our personalized FL approach for pain estimation in a real-world setting (e.g., using mobile phones to monitor pain) is shown in Fig.~\ref{fig:federated-personalization}. We depict there the communication rounds during the training of the deep model for confidentiality-preserving pain estimation. To facilitate further evaluation of the models introduced in this paper and encourage future research in this important direction, the code used to conduct the experiments in this paper is made publicly available and can be downloaded from \url{https://github.com/ntobis/federated-machine-learning}.

\section{Background Work}
\subsection{Federated Machine Learning}

Federated Machine Learning was first introduced by Google in 2016~\cite{ref-google-FML-init,ref-google-FML-init2,ref-google-FML-init3}. Different from a centralized setting, in a FL setting, multiple devices e.g., end-user devices such as mobile phones, or business infrastructure such as hospital servers, contribute to learning a machine learning classifier. The classifier can be a deep neural network, but also can be a simpler model with fewer parameters such as a support vector machine, or a logistic regression model~\cite{federated-log-regression}. FL models are distinct in that the original training data never leaves the respective local device that collected it. Each device (also dubbed \textit{client}) maintains a version of the same model, which is updated with every new observation. The updates to the model (i.e., the updated weights and biases of neurons in a deep network), and not the observations (in our case, the face images) themselves, are then shared with a central server, which averages the parameters of the newly trained models from all participating devices. Once a new version of the model has been trained, it is pushed back down to all clients. This process repeats continuously until the model converges, and/or the training data is exhausted. The key difference of the personalized FL approach introduced in this paper is that the standard FL performs averaging of all the model parameters - by contrast, in our approach the last (decision) layers of the deep models are not shared and are updated only locally. In this way, we enable the models to easily personalize to target clients while introducing another level of confidentiality-preserving (since the parameters of the last layer of our deep models for pain estimation are never shared with a central server, see Fig.~\ref{fig:federated-personalization}). 

The approach proposed here is also related to the recently proposed concept of {\it split-learning} (\url{https://splitlearning.github.io}), where only parts of the model, i.e., a subset of its parameters are shared across different clients (in their case, different hospitals)~\cite{poirot2019split}. However, in contrast to our approach, their model  'splitting' is intriguingly a kind of opposite of what our approach is doing - their {\it split-learning} is holding back the early layers of the deep model, not the final layer, as proposed here - which, in the context of pain estimation, is critical for enabling efficient model personalization. 

\subsection{Applications of Federated Learning}
Potential applications for FL are vast and differ substantially in vertical and specific use-case, but they typically bear three common traits~\cite{ref-google-keyboard}: (i) Task labels do not necessarily need to be provided by humans but can be derived naturally from user interaction. (ii) Training data are confidentiality-sensitive, and (iii) Training datasets are large, and hard to collect in a central location. Not all of these conditions strictly need to hold when applying FL, but it is under these circumstances that FL tends to provide the most value over other machine learning algorithms. To illustrate the potential of FL, we briefly review some recent applications of FL models.

For instance, the computer chip-maker Intel \cite{fed-intel} leverages FL to showcase how multiple healthcare institutions can collaborate in a confidentiality-preserving manner, leveraging each institution's electronic health records (EHR). The authors argue that while collaboration between institutions could address the challenge of acquiring sufficient data to train machine learning classifiers, the sharing of medical data is heavily regulated and restricted. They present the first use of an FL classifier for multi-institutional collaboration and find that they can learn a federated semantic model for brain tumour segmentation that compares favorably to a model trained on centralized data. Further evidence of benefits of leveraging FL from the health records under data-privacy constraints can be found in recently published works \cite{brisimi2018federated,choudhury2019differential,liu2018fadl,choudhury2019predicting}. These support the hypothesis that FL can lead to increased data-privacy protection within the health space, which is also discussed in more detail in a recent overview of digital health applications with FL \cite{Rieke2020TheFO}.  

Beyond health, FL has also been used in other applications such as using data from 360K users to improve the search results in the Firefox Search Bar \cite{hartmann2018federated}, without collecting the users' actual data. Millions of URLs are entered into Firefox daily, notably improving the auto-complete feature for users, while enhancing user experience and potentially increasing retention. Likewise, Google describes one of the first implementations of FL on a large scale, training a global model "to improve virtual keyboard search suggestion quality"~\cite{ref-google-keyboard}. In this work, the authors address the technical challenges of coordinating the model training on millions of devices worldwide. Examples include connectivity issues, the bias of training a model across different time zones, and minimizing the impact on user experience that training a machine learning model locally has (e.g., battery-life and device-speed). They note that future work on privacy still needs to be done and cautiously call their method "privacy-advantaged", vs. "privacy-preserving". For similar reasons, in this work we adopted the term "confidentiality-preserving" instead of "privacy-preserving".

The works mentioned above demonstrate empirically the benefits of FL on real-world problems; however, to the best of our knowledge, the FL approach has not been explored before in the context of pain estimation from face images. This is a challenging problem where data are non-stationary, i.e., they change over time (the same person may show facial expressions of different pain levels in different recording sessions). Furthermore, the face-image data of positive and negative examples of pain (i.e., pain vs. no pain) are highly imbalanced and not easily distinguishable as the underlying data distribution also evolves over time and space (different camera orientations, lighting, and so on), as would be expected in real-world applications. Taken together, all these factors make the problem of pain estimation from the real-world face images highly challenging, which is even more true in the case of distributed learning algorithms such as FL.

\subsection{Practical Challenges for Federated Learning}
As with the standard machine learning, FL is also sensitive to some of the data properties. Below we emphasize the key aspects of the data that directly affect the performance of the FL algorithms. for more detailed summary of challenges and open problems in FL, see~\cite{kairouz2019advances}.

\begin{figure}
    \centering
    \includegraphics[width=.40\textwidth]{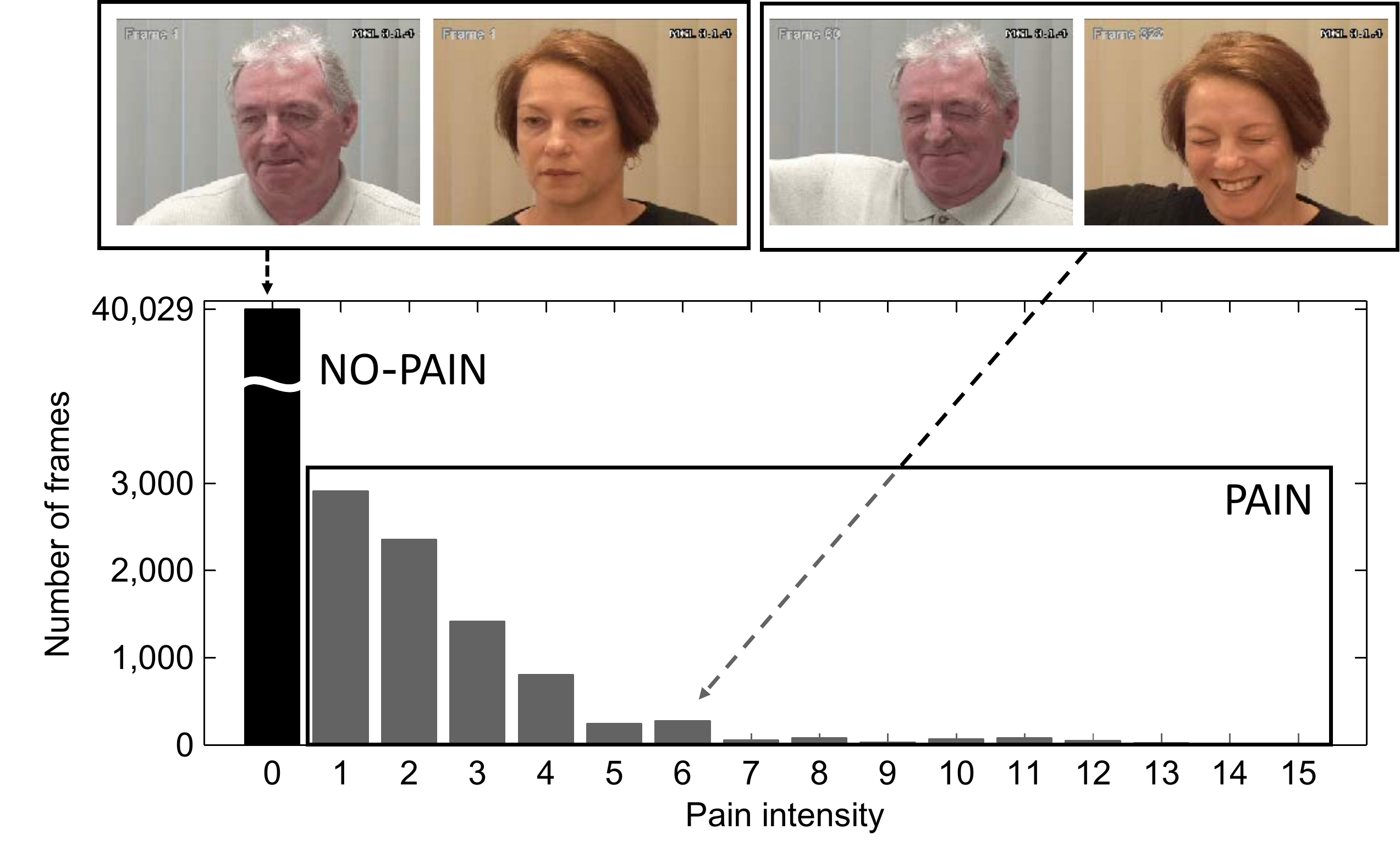}
    \caption{Distribution of pain levels (PSPI) in the the UNBC-McMaster dataset~\cite{painful-data}.}
    \label{fig:painlevels}
\end{figure}

\paragraph{Non-IID and Unbalanced Data} If we draw a sufficiently large sample \textit{at random} from the overall data distribution, we can state, with a specified level of confidence, that the sample is representative of the overall population (independent and identically distributed - IID). In FL, clients' datasets will often differ substantially from those of other clients (e.g., in the case of mobile phones the phone's dataset is dependent mostly on the interaction with one particular user). Specifically, in the case of pain estimation from face images, there are large individual differences in the face morphology and facial texture, dynamics of facial expression and also expressiveness of pain (some subjects have their facial expressions of pain toned-down due to long-term exposure to pain) -- see Fig.~\ref{fig:painlevels}. Moreover, the factors such as the camera view, illumination changes, and similar contribute significantly toward violating the IID assumption. Thus, sampling a client's data at random is not likely to yield training data that are representative of the target population, making FL sensitive to the artifacts mentioned above. 

Furthermore, the clients' data may vary substantially in size and diversity. As mentioned above, in the dataset for pain estimation used in this work, some subjects have considerably less instances of "pain" images than the others. Moreover, for all the subjects, the data labels are highly dominated by the negative class (no pain), which can be observed from Table.~\ref{tab:sessions} on the individual level, and overall in Fig.~\ref{fig:painlevels}. Also, since we are dealing with the real-world longitudinal recordings of each subject (i.e., different recording sessions), some subjects may or may not participate in certain sessions, and a majority of the subjects have data for only the first few sessions (see Table.~\ref{tab:sessions}). While this creates a highly challenging learning context for both the centralized learning and FL, it is more pronounced in the latter case as we need to ensure that {\it each individual} (instead of the whole group as in case of centralized learning) provides enough and balanced data for the local (e.g., on-device) fine-tuning of the federated models. Since the global models are derived through averaging of the local model parameters, it is thus important for FL to have access to robust locally learned models prior to the averaging of their parameters. Therefore, any corrupted local-models (trained on highly imbalanced data) can adversely affect the global model parameters. 

\paragraph{Limited Communication} Mobile phones are frequently offline or are connected to flaky or expensive internet connections. Healthcare facilities, especially in rural areas of developed countries such as the United States, often have slow internet connections or a minimal number of computers that are linked to the internet \cite{broadband}. While it is usually cheap to compute the updates locally, when the amount of training data is small, communicating these local updates becomes much more time-consuming, making communication-speed and averaging the main bottleneck in FL. In this work we use data of naturalistic facial expressions of pain by different subjects, recorded at different physiotherapy sessions, to simulate the real-world FL of their models for pain estimation. Note however that in this dataset not all the subjects are present in all the recording sessions, and/or may have short sessions, sometimes without the pain expressions (the target exercise may not induce pain at the specific times). While this is likely to adversely affect FL of the deep models for pain estimation, we use such data to evaluate the robustness of our algorithms under these challenging conditions.

\paragraph{Maintaining performance} Another important aspect of FL is that the global model (obtained by combining the parameters of the federated models) is not trained on the raw data (face images) -- it is trained using {\it only} the parameters of the locally-tuned models (for each client), as only the local models have access to the raw data (face images and their labels). Thus, unconventional learning strategies are needed to achieve similar performance in a federated setting compared to a centralized machine learning approach. If a model fully preserves the confidentiality of a client but produces inaccurate predictions, it is essentially useless. Therefore, to start the learning of our FL models, we assume that we are given an initial global model trained on a sub-population of patients (e.g. those who opted in to share their face data). Once such model is instantiated and learned centrally, we use it as a starting point to perform further learning via FL. During this time, the goal of the learning is to produce model parameters that would enable better generalization and no regression in the performance, in terms of pain estimation, for the target subjects.\newline

In summary, the challenges of FL that we described above are all important and pose a serious bottleneck for FL algorithms. In this work, we tackle the challenge of learning from non-IID and unbalanced data, by applying data augmentation via targeted image pre-processing, and class-stratified sampling of the client's data, to mitigate those challenges (see Sec.~\ref{painEstimation} for more details). For the challenges associated with the limited communication and performance maintenance, we design a novel training and evaluation scheme that enables to learn the model parameters from one-session-at-the-time of the target client (thus being adjusted for the longitudinal nature of the data used in this work). Moreover, to prevent the effect of catastrophic forgetting when training neural networks~\cite{kirkpatrick2017overcoming}, we use the new session data of the target client to validate instead of to train the local models (later those data are used to train the model as well when a new round of the FL communication starts). For details of the newly introduced model training approach, see Fig.\ref{fig:sessions}.  

\subsection{Automated Pain Estimation}
\label{painEstimation}
The importance of regularly checking on a patient's well-being is described in Atul Gawande's "The Checklist Manifesto" ~\cite{checklist-manifesto}. While there is evidence suggesting that improved pain monitoring (either by the nursing staff or through self-reports) yields better patient outcomes~\cite{Wells_chapter17}, such an approach is often difficult to implement in practice. This is because of the competing demand for nursing staff~\cite{economist-nursing}, and the need for patients to regularly provide their self-report of pain intensity (e.g., using the Visual Analog Scale (VAS), a subjective measure for acute and chronic pain, commonly used to rate patients' pain level on a scale 0-10).

Automated pain monitoring from face images could provide a more objective and {\it continuous} (in time) measurement of pain, and also help to differentiate real from fake pain in order to provide adequate dosage of medication for treating the underlying condition. According to Prakchin and Solomon~\cite{prkachin1992consistency}, the bulk of what humans experience as pain can be described by the activations of six (out of 44) facial Action Units (AUs), defined by the Facial Action Coding System (FACS)~\cite{ekman2002facial}. These include: brow lowering (AU4), orbital tightening (AU6 and AU7), elevator contraction (AU9 and AU10) and eye closure (AU43). A follow-up paper presents the Prkachin and Solomon Pain Intensity (PSPI) metric \cite{prkachin2008structure}, proposing the level of pain as the following function of facial AU intensities:
\begin{equation}
    Pain = AU4 + max(AU6, AU7) + max(AU9, AU10) + AU43.
\end{equation}
The result is a 16-point scale, where the first five AUs are scored on a 6-point scale (0 = absent to 5 = maximum intensity), and the activation of AU43 (eyes open/closed) is binary. For an example of pain vs. no pain face images, see Fig.~\ref{fig:painlevels}. To obtain the PSPI pain labels, the face images are manually coded in terms of AUs by FACS~\cite{ekman2002facial} certified experts. This manual process is highly tedious, time-consuming, and costly. To this end, a number of methods for automated pain estimation from face images have been proposed (for surveys, see~\cite{aung2015automatic,chen2018automated, hammal2018automatic, werner2019automatic}). 

Due to subtle changes in facial expressions and inter-subject variability, per frame PSPI estimation is very challenging. For this reason, many researchers have attempted to automate pain detection (pain vs. no pain) instead of the full-range intensity estimation of pain, for instance, using Active Appearance Model (AAM) - based features combined with a Support Vector Machine (SVM) to classify pain versus no pain images~\cite{painful-data}. Others ~\cite{hammal2012pain} have addressed estimation of pain intensity on a 4 level scale, using one-versus-all SVM classifiers, while others ~\cite{kaltwang2012continuous} have used Relevance Vector Regression to estimate the full pain range (0-15). Some have used temporal models for pain estimation, e.g., using long short-term memory (LSTM) neural networks~\cite{rodriguez2017deep}. In this approach, raw face images are directly fed to a CNN that extracts the frame-based features. These features are then used as input to LSTMs to predict the PSPI score per frame. A number of other deep-learning models have recently been proposed for pain estimation (e.g.,see ~\cite{egede2017fusing, facial-action-unit,soar2018deep}).   

In this work, we focus on pain estimation per frame and in terms of binary detection of pain (0 - no pain, 1 - pain) because of the highly imbalanced nature of the pain PSPI scores. Namely, we group all the face images with PSPI 1-15 as pain, and those with PSPI=0 as no pain. While this reduces the label imbalance, even the binary pain labels are highly imbalanced per subject/session in the dataset used (see Fig.~\ref{fig:painlevels}). This already poses a serious challenge for FL from different subjects.

\section{Methods}\label{methods}
\subsection{Data Pre-Processing and Augmentation}
\label{augmentation}
Each image frame of the 200 video sequences from the UNBC database is a color image of 250 x 250 pixels. As color is not relevant to detecting pain using the FACS system, we greyscaled all images, to reduce the number of input features required to pass to the network by two thirds. Following this step, we performed histogram equalization for each image to increase its contrast. Since the Prkachin and Solomon Pain Intensity Scale is measured by looking at only a limited number of features on a person's face, we want the appearance of these features to be as poignant as possible. Increasing the image's contrast makes features like the person's eyes or eyebrows stand out further.

As can be seen from Fig.~\ref{fig:painlevels}, examples of test subjects experiencing no pain significantly outnumber those experiencing pain. This imbalance would also be expected in a `real-world' example, where patients may only experience pain sporadically during their hospital visit. We upsample the positive examples by applying two data augmentation techniques: we first created a flipped copy of all images. We then created another copy of all originals and flipped copies, respectively, that was randomly rotated by either 10 degrees to the left or the right. After the rotation, each image was cropped from 250 x 250 pixels to 215 x 215 pixels in order not to have any whitespace or artificial filling around the images. To make the input shape consistent across mutations, all images were cropped to 215 x 215 pixels. We underscore the importance of applying these steps to both positive and negative examples, as the network might otherwise learn that a specific image mutation always represents a positive example. Finally, to balance the dataset, we downsampled the majority class (no-pain) samples so that it matches the number of the train pain images, i.e., of the minority class (see Sec.~\ref{experimental-setting} for more details). 

\subsection{Deep Learning for Pain Estimation}
\begin{figure}
    \centering
    \includegraphics[width=.45\textwidth]{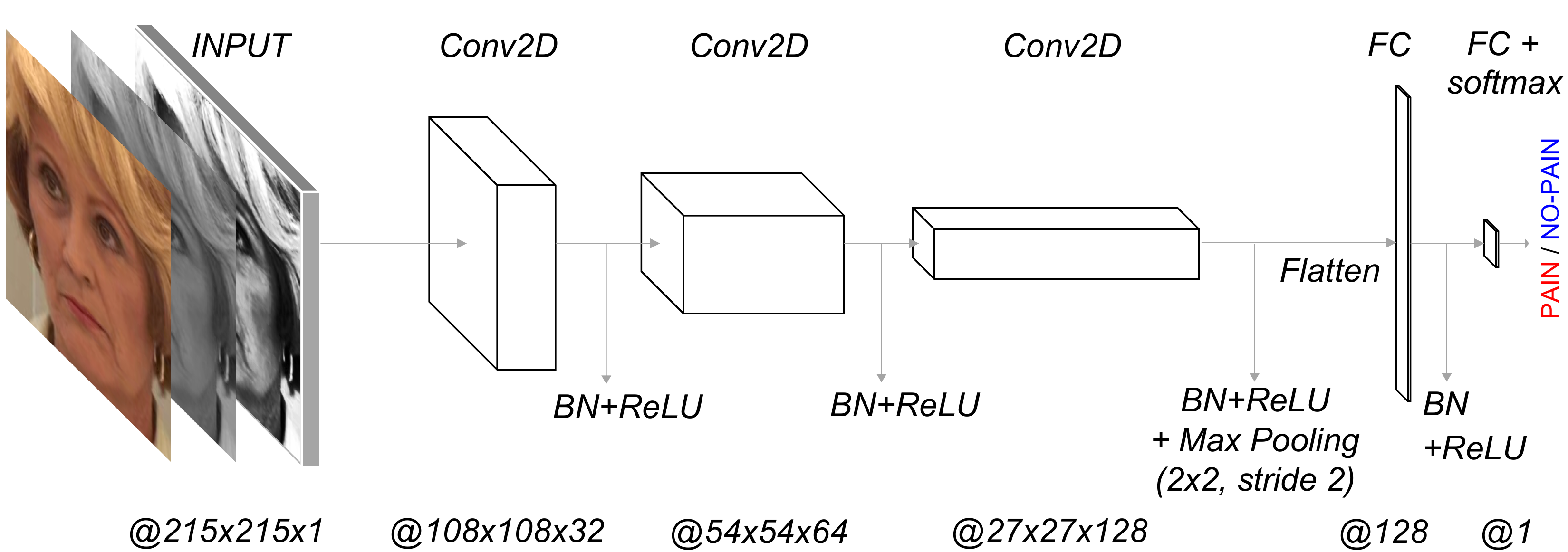}
    \caption{Network architecture of the base deep model for pain estimation.}
    \label{fig:deepmodel}
\end{figure}

In part inspired by \textit{MobileNets: Efficient Convolutional Neural Networks (CNNs) for Mobile Vision Applications}~\cite{mobilenets}, we designed a lightweight CNN-based model architecture (see Fig.~\ref{fig:deepmodel}) that can also be trained for a small number of epochs on hardware-limited computing devices. Similar architectures have previously shown success in the task of facial expression recognition from face images, including facial expressions of pain~\cite{facial-action-unit}. 

Our base deep model for pain estimation takes as input a (215, 215) re-scaled and processed face image, as explained above. This representation is then fed into 3 convolutional layers for feature extraction. The first layer has 32 filters, a kernel size of (5x5), and is followed by a batch normalization (BN) and ReLU layer. The convolutional layers are followed by 2x2 max-pooling (with stride 2), and two fully-connected layers: with 128 units, and  with 1 unit plus the sigmoid activation (see Fig.~\ref{fig:deepmodel}). We used the SGD optimizer with a learning rate of $10^{-4}$, and the binary cross-entropy loss function. We applied early stopping when there was no improvement in the loss function after 5 training epochs. This helped to reduce overfitting. We also experimented with the well-known pre-trained deep architectures such as ResNet50 \cite{resnet} and VGGNet \cite{vgg} by fine-tuning the last two layers of the networks. However, as also found in~\cite{facial-action-unit}, for the target task, simpler network architectures work better in practice. Therefore, we used the architecture from Fig.~\ref{fig:deepmodel}.

\subsection{Federated Learning}
\label{fedlearn}
In FL, we assume that the dataset is not centrally stored, but is partitioned over  $k = 1, \ldots, K$ clients. We use the following notation:
\begin{itemize}
    \item Each partition is represented as a set of indices $\mathcal{P}_{k}$ of data points that a given client $k$ holds
    \item $n$ represents the number of all data points collected by all clients and thus $n_k$ represents the number of data points that the client holds
    \item $n_k = |\mathcal{P}_k|$.
\end{itemize}

If the standard definition of minimizing a loss function is given by:
\begin{equation}
    \min_{\theta \in \mathbb{R}^d} f(\theta), \,\, {\text{where}} \,\, f(\theta) \stackrel{def}{=} \frac{1}{n} \sum_{i=1}^{n} f_i(\theta),
\end{equation}
and $f_i(\theta)$ represents the loss for a prediction of one observation $(x_i, y_i)$ given model parameters $\theta$, this loss function can be rewritten to represent $K$ clients in a federated setting, such that:
\begin{equation}
    f(\theta) = \sum_{k=1}^{K}\frac{n_k}{n}F_k(\theta),
\end{equation}
where:
\begin{equation}
    F_k(\theta) = \frac{1}{n_k}\sum_{i\in\mathcal{P}_{k}}f_i(\theta).
\end{equation}

Thus, instead of computing loss (i.e., the binary cross-entropy) as an average over $n$ number of samples from a centralized data set as: $\frac{1}{n}\sum_{i=1}^{n}f_i(\theta)$, we compute the average loss $F_k(\theta)$ for a specified client $k$ as $\frac{1}{n_k}\sum_{i\in\mathcal{P}_{k}}f_i(\theta)$ and then group the loss of all participating clients $K$, by computing a weighted average loss based on the number of labelled data points $n_k$ that each client holds.\newline

Similar to computing the loss, we also compute the gradients of the federated model. In a federated setting each client computes the average gradient $g_k$ on its local data as:
\begin{equation}
    g_k = \nabla F_k (\theta)\,.
\end{equation}
Then, each client takes a step of gradient descent and updates its parameters accordingly, formalized as:
\begin{equation}
    \forall k, \theta_k \leftarrow \theta - \eta g_k\,.
\end{equation}
This step can be repeated multiple times, i.e., for multiple epochs $E$, until a central server then computes the weighted average of these gradients similar to the weighted average of the loss above as:
\begin{equation}
    \theta \leftarrow \sum_{k=1}^{K} \frac{n_k}{n}\theta_k,
\end{equation}
to update the model parameters of the overall model, stored on the central server. This concludes a full round of updates to the global model.\newline

Assuming mini-batch stochastic gradient descent, in such a federated setting the computational effort for one full update is controlled by three parameters:
\begin{enumerate}
    \item The fraction $C$ of clients $K$ that participate in a given update round.
    \item The number of steps of gradient descent (or epochs) $E$ that each client performs.
    \item The batch size $B$ used for all client updates.
\end{enumerate}
While $C$ impacts the computational power required at the server level (more participating clients requires more transfer of data to the server and more effort in aggregating information), $E$ and $B$ impact the computational effort required on the client-side. The complete pseudo-code for this approach was first proposed in \cite{ref-google-FML-init}.

\begin{algorithm}[ht]
\caption{\texttt{Federated Personalization}. The \textit{K} clients are indexed by \textit{k}; \textit{B} is the local minibatch size, \textit{E} is the number of local epochs, \textit{F} denotes the rounds that the local model is fine-tuned, and $\eta$ is the learning rate. $\alpha$ denotes the factor by which the learning rate is decreased in the \texttt{ClientFineTuning} procedure. $w_g$ denotes the global model parameters (the shared layers), $w_l^{k=1,\dots,K}$ denotes the local model parameters (only the prediction layer), and $\theta=\{w_g \cup w_l^{k=1,\dots,K}\}$.}
\label{alg:personalized-learning}
\begin{algorithmic}[0]
\Procedure{Server executes:}{}
    \State initialize $w_{g_0}$
    \State initialize $w_{l_0}$
    \For{each round $t=1,2,...,T$}
        \State $m \leftarrow$ max($C \times K,1$)
        \State $S_t \leftarrow$ \text{(random set of} $m$ \text{clients)}
        \For{each client k $\in S_t$ \textbf{in parallel}}
            \State $w_{g_{t+1}}^k \leftarrow$ \textproc{ClientUpdate}($w_{g_t},w_{l_t}^k$)
        \EndFor
        \State $w_{g_{t+1}} \leftarrow \sum_{k=1}^K \frac{n_k}{n}w_{g_{t+1}}^k$
        \For{each client k $\in S_t$ \textbf{in parallel}}
            \State \textproc{ClientFineTuning}($w_{g_{t+1}},w_{l_{t+1}}^k$)
        \EndFor
        \State $\theta_{t+1}=\{w_{g_{t+1}} \cup w_{l_{t+1}}^{k=1,\dots,K}\}$ used for inference in this round
    \EndFor
\EndProcedure
\State
\Procedure{ClientUpdate}{$w_g, w_l$} \Comment{Run on client k}
    \State $\mathcal{B} \leftarrow$ (split $\mathcal{P}_k$ into batches of size $B$)
    \For{each local epoch $i$ from 1 to $E$}
        \State $\{w_g \cup w_l\} \leftarrow \{w_g \cup w_l\} - \eta\nabla\ell(\{w_g \cup w_l\};b)$ \Comment{Update all layers}
    \EndFor
    \State return $w_g$ to server
\EndProcedure
\State
\Procedure{ClientFineTuning}{$w_g, w_l$} \Comment{Run on client k}
    \State $\mathcal{B} \leftarrow$ (split $\mathcal{P}_k$ into batches of size $B$)
    \For{each local epoch $i$ from 1 to $F$}
        \State $w_l \leftarrow w_l - \alpha\eta\nabla\ell(w_l;b)$\Comment{Update local layers only}
    \EndFor
\EndProcedure

\end{algorithmic}
\end{algorithm}
\subsection{Federated Personalization}
\label{sec:fl}
Our benchmark FL algorithm mirrors the \texttt{FederatedAveraging} algorithm proposed in \cite{ref-google-FML-init}, originally applied to the Google Keyboard, to suggest improvements to the next iteration of Gboard’s query suggestion model for each user's device. For our purposes, each target subject from the UNBC dataset represented one user. In this context, the underlying premise of FL is that instead of performing the centralized learning of the models, the models can be learned locally and by averaging the parameters across the clients, the resulting parameters performing similarly to centralized learning. For theoretical justification of this, see~\cite{ref-google-FML-init,bonawitz2019towards}.
In the proposed FPDL approach, only the weights of the convolutional layers (no feature maps) are sent to the central server for averaging, while the weights of the fully-dense output layers are retained locally (i.e., their parameters are never shared). This is for two reasons. First, \cite{hitaj2017deep,melis2018exploiting} showed that even when only the network parameters are shared with a central server, an adversary can use a generative adversarial network (GANs)~\cite{goodfellow2014generative}, to iteratively learn new pieces of information about another client's training data, potentially becoming able to reconstruct private information about the client (in our case, the face images\footnote{It would still be very difficult to reconstruct the high-dimensional face images, given that the CNN feature maps are not shared during federated learning in our approach.}). Without access to all layers in the network, an adversary misses a critical building block to infer the subject's pain level. This results in an additional layer of confidentiality to the traditional  \texttt{FederatedAveraging} algorithm.\par

Second, applying the FL optimization algorithm to the lower convolutional layers enables the model to learn general facial features from a large population. On the other hand, the final upper layers of a deep network are mostly informative for the final prediction task (pain/no pain). Thus, they do not necessarily need to learn from the entire population, and could potentially even benefit from learning only the target subject's pain expressions, as subtle facial expressions might only generalize well for that subject, but not for the whole population. This intuition is the basis of our FPDL algorithm, described in detail in Alg.1, and depicted in Fig.~\ref{fig:federated-personalization}.

To start, all clients receive a model whose parameters are initialized with the pre-trained parameters of a model that was trained centrally, without personalization, on the same task, with a different subset of the data. The algorithm then follows the steps of the \texttt{FederatedAveraging} algorithm, except that once local training is complete, i.e. procedure \texttt{ClientUpdate} concludes, \textbf{only} the weights of the convolutional layers $w_g$ are sent to the central server for averaging. The weights of the final full-connected layers $w_l$ stay locally with the client $k$, just like the data. Consequently, only the convolutional weights are averaged to $w_{g_{t+1}}$ and then used to update each local client model. We must then engage in "local fine-tuning" - see procedure \texttt{ClientFineTuning}. To fine-tune, we "freeze" the convolutional layers $w_g$ for each client and decrease the optimizer's learning rate $\eta$ for each client by a factor $\alpha$ in order to avoid overshooting. The decrease in learning rate ensures that the local layers slowly "reconnect" to the previously detached, globally averaged, convolutional layers. We then train the local models for a number of epochs, before using the resulting $\theta_{t+1}=\{w_{g_{t+1}} \cup w_{l_{t+1}}^{k=1,\dots,K}\}$ for inference on new data. The final \texttt{ClientFineTuning} step ensures that models become personalized, as each client's dense layers are only trained on locally available data during this step. We implemented this approach in Python using the Keras framework and the tensorflow back-end; the code is made publicly available and can be downloaded from \url{https://github.com/ntobis/federated-machine-learning}.

\section{Experimental Setting}\label{experimental-setting}
\label{dataset}
\paragraph{Data.} To evaluate the models, we used the UNBC-McMaster Shoulder Pain Expression Archive database~\cite{painful-data}, containing 48,106 video frames from 25 test subjects undergoing a range-of-motion tests to their affected and unaffected limbs. These video frames come from up to 10 sessions per subject, and are annotated using the PSPI scores, as described in Sec.~\ref{painEstimation}. Since these pain labels are highly imbalanced, we categorized every example of level 1 or higher as {\it pain}, and categorized level 0 as {\it no pain}. The summary of the label distribution per test subject/session is reported in Table~\ref{tab:sessions}. 

\begin{table}[ht]
    \centering
    \resizebox{\columnwidth}{!}{%
    \begin{tabular}{rrrrrrrrrrrrrr}
    \toprule
           & \multicolumn{10}{c}{\textbf{Positive Examples per Session}} & \multicolumn{3}{c}{\textbf{Total}} \\
    ID &       0 &    1 &    2 &    3 &    4 &    5 &    6 & 7 &    8 &    9  &   Pain & No Pain & Pain \% \\
    \midrule
        43 &     140 &      &      &      &  228 &      &      &   &      &       &    368 &   4,112 &     8\% \\
        48 &         &  148 &      &      &      &  188 &      &   &      &       &    336 &   3,192 &    10\% \\
        52 &      72 &      &      &      &      &      &   44 &   &  120 &  188  &    424 &  10,012 &     4\% \\
        59 &         &  532 &      &      &      &      &      &   &      &       &    532 &   2,560 &    17\% \\
        64 &     244 &   64 &   64 &      &  248 &      &      &   &      &       &    620 &   5,576 &    10\% \\
        80 &   1,052 &  536 &  484 &  484 &  660 &  792 &  264 &   &      &       &  4,272 &   3,584 &    54\% \\
        92 &     464 &  696 &      &      &  724 &      &      &   &      &       &  1,884 &   4,124 &    31\% \\
        96 &         &      &      &      &  112 &      &  512 &   &   88 &       &    712 &   8,700 &     8\% \\
       107 &      32 &      &  848 &   60 &  828 &      &      &   &      &       &  1,768 &   6,396 &    22\% \\
       109 &         &  600 &      &      &      &  116 &      &   &      &      &    716 &   6,896 &     9\% \\
       115 &      60 &      &  220 &   56 &   60 &      &      &   &      &       &    396 &   4,736 &     8\% \\
       120 &     116 &      &      &  188 &      &      &      &   &      &       &    304 &   5,960 &     5\% \\
    \bottomrule
    \end{tabular}%
    }
    \caption{Number of positive examples by session and target subjects. Each target subject participated in as many consecutive sessions as specified in the column \textit{\# of Sessions}, starting with session 0. No number indicates no positive examples for that session (but negative examples, if the session index is smaller than \textit{\# of Sessions}). We also show the distribution of the pain/no pain labels per target subject after binarizing the 16 pain levels (0) - no pain, (1-15) - pain, from the UNBC-McMaster dataset. Note that the resulting distribution is still highly imbalanced for most subjects.}
    \label{tab:sessions}
\end{table}

\begin{figure}[ht]
    \centering
    \includegraphics[width=0.45\textwidth]{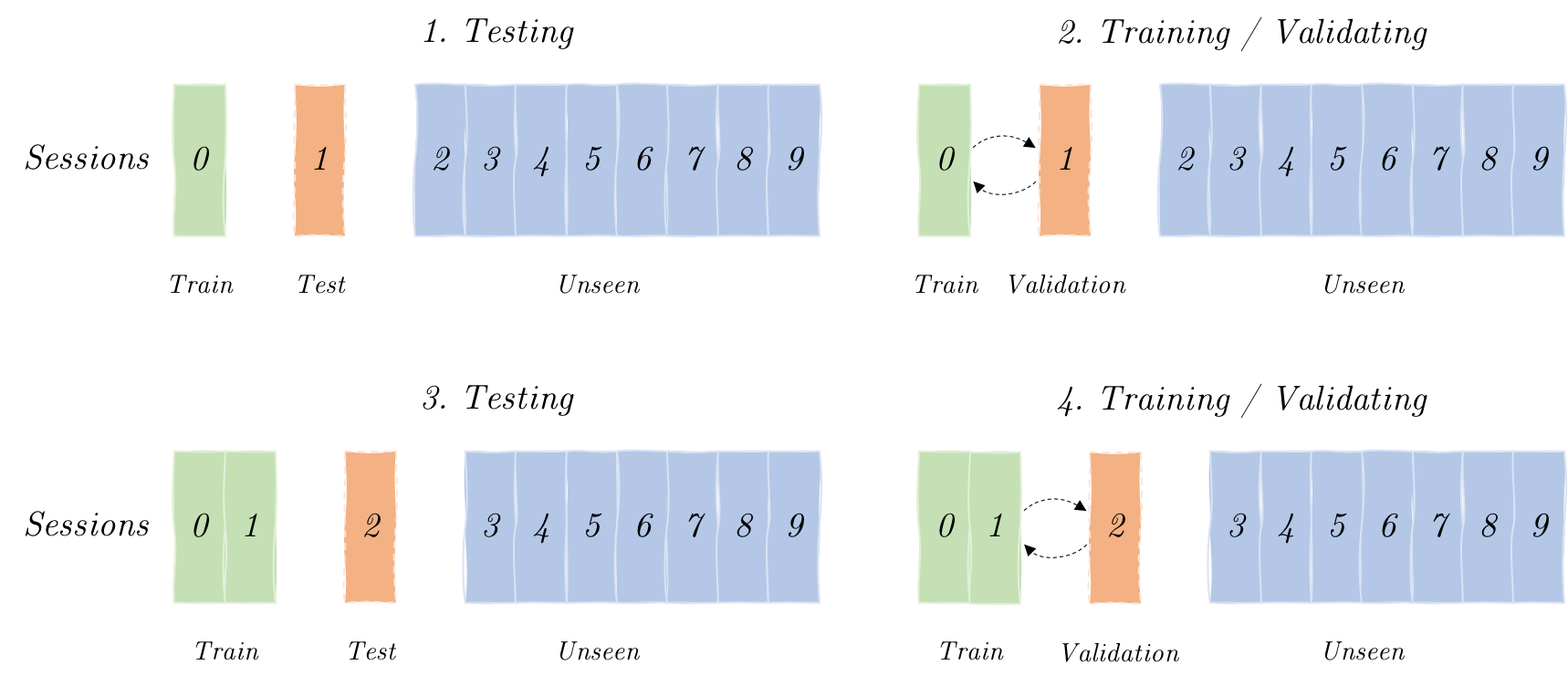}
    \caption{The training/validation/test protocol. The first session is zero-indexed. In step (1), the pre-trained model is tested on session 1. In step (2), the model is trained on session 0, using session 1 as a validation set to apply early stopping. In step (3), the model trained in (2) is tested on session 2. In step (4), session 1 has become part of the training data, i.e., the model is then trained on sessions 0/1, and validated on session 2. This process is repeated until the last session.}
    \label{fig:sessions}
\end{figure}

\paragraph{Evaluation Setting}\label{evaluation-setting} The models from Sec.~\ref{methods} were first pre-trained on data of a subset of 13 subjects. The further model learning and evaluation was then performed on the remaining 12 subjects ("test subjects"). To simulate a real-world setting, we assumed that each session only becomes available once the model has been trained on the previous session. The new session first serves as an unseen test-set, and is then converted into a validation set, and is finally merged into the training data, once a new session becomes available (see Fig.~\ref{fig:sessions} for details). Therefore, no data from the current (test) session is used to train the models. To balance the intra-session data for each client we sampled at random 200 positive/negative pain examples, with replacement, for each subject, from each session that was available for that subject, leading to 400 images per target subject/session (see also Sec.~\ref{painEstimation}). Using this data balancing strategy and also the data from the previous session to validate the model parameters was important to achieve robust FL of the deep models. \par

\paragraph{Models.} The models compared are: the base CNN (Sec.~\ref{methods}) trained in a centralized manner on non-target subjects, i.e., the subjects used to pre-train the deep model. We refer to this model as the base centralized deep learning (BCDL), serving as the baseline. We also perform centralized deep learning on target subjects (CDL), as well as the traditional Federated Deep Learning (FDL), described in Sec.~\ref{fedlearn}. For comparisons, we also include subject-specific models trained locally using data of the target subjects only (LDL). Note that all of the learning algorithms use the base CNN as their underlying model, and 
the training of the CDL, FDL and LDL models is performed using the same protocol as described in Fig.~\ref{fig:sessions}. We also report the results by the randomly initialized CNN model (RND), where no model training was performed. We ran all of the models with 10 random seeds, and report the average results.
\begin{table*}
    \centering
    \resizebox{.7\textwidth}{!}{%
    \begin{tabular}{>{\bfseries}p{2.5cm}rrrrrrrrrrrrc}
    \toprule
     \textbf{Method \,\,\,\,\,\,| \,\,\,\,\,ID} & \textbf{\#43} &  \textbf{\#48} &  \textbf{\#52} &  \textbf{\#59} &  \textbf{\#64} &  \textbf{\#80} &  \textbf{\#92} &  \textbf{\#96} &  \textbf{\#107} &  \textbf{\#109} &  \textbf{\#115} &  \textbf{\#120} &  \textbf{Average ($\pm$std)} \\
    \midrule
           RND &  35 &  \textbf{24} &  10 &  37 &  14 &  44 &  51 &  25 &   42 &   31 &   15 &   36 &        30 $\pm$ 13 \\
           BCDL &  24 &   5 &  32 &  \textbf{56} &  25 &  \textbf{60} &  79 &   8 &   59 &   30 &   35 &   \textbf{65} &       40 $\pm$ 24 \\ \hline
            CDL &  64 &   7 &  30 &  \textbf{56} &  \textbf{39} &  53 &  67 &  45 &   62 &   32 &   46 &   61 &       46 $\pm$ 18 \\
      {\text LDL} &  \textbf{75} &   7 &  35 &  \textbf{56} &  34 &  49 &  53 &  37 &   \textbf{68} &   33 &   42 &   46 &       44 $\pm$ 19\\ 
      FDL &  27 &   7 &  \textbf{44} &  \textbf{56} &  25 &  58 &  \textbf{81} &  \textbf{46} &   63 &   \textbf{42} &   35 &   62 &       46 $\pm$ 20 \\ \hline
     {\bf PFDL} &  72 &   7 &  36 &  \textbf{56} &  37 &  52 &  76 &  35 &   \textbf{68} &   29 &   \textbf{47} &   48 &       \textbf{47 $\pm$ 20} \\
    \bottomrule
    \end{tabular}
    }
    \caption{F1 score (in \% and rounded to the nearest integer) per target subject. Best model for each subject is highlighted in bold. The results are averaged across the available sessions of target subjects.}
    \label{tab:test-subject-results}
\end{table*}
\section{Results}
Table~\ref{tab:test-subject-results} shows the pain estimation performance per subject (averaged across available sessions for that subject) in terms of F-1 score. First, we observe that all models outperform the random approach, as expected. Note that this approach achieves an F-1 of $30\%$, instead of $50\%$, which would be the chance level for a binary classification. This is because of the unbalanced test data. The pre-trained centralized model (BCDL) performs worse, on average, than the models that were updated across the sessions using the target subjects {\it past} data (see Fig.~\ref{fig:sessions}). 

Specifically, the centralized deep model (CDL) gains $6\%$ in performance due to the adaptation -- the same performance achieved by the standard federated deep learning (FDL). This result supports the robustness of the proposed training strategy, which reaches the same performance without sharing the raw image data of each target subject but only the models' parameters. Likewise, the subject-specific models (LDL), trained using only the data of the target subject also have improvements over the pre-trained BCDL model. However, they do not reach the full performance of the CDL and FDL models, which leverage the data/parameters of the other subjects. On the other hand, the personalized federated model (PFDL) reaches the performance of the fully federated model (FDL), even outperforming it slightly on average ($1\%$). This shows that PFDL can retain the performance of the fully federated or centralized-learning models while adding an additional layer of confidentiality, i.e., without sharing the parameters of the final personalized classification layer.

From the performance of the models at the individual level, we make the following observations. First, note that for subject \#43, both LDL and PFDL significantly outperform both CDL and FDL. By inspecting the data of this subject in Table~\ref{tab:sessions}, we see that only two sessions were available: thus, PFDL enabled learning of the local models better than through the standard averaging across sessions as done by FDL. Similar conclusions can be derived for subjects \#64 and \#107. On the other hand, for subjects where FDL outperforms the other models, in most cases the performance of PFDL is close and better than that of the local LDL models. Thus, we note a general trend of the PFDL model: it provides a good trade-off between the subject-specific models and fully federated models, likely because it performs subject-specific adaptation of the last (classification) layer while always taking into account the data of other subjects, rendering it more robust in this challenging scenario of learning from highly unbalanced data. 

We also note that all of the models fail on subject \#48. Looking at Table~\ref{tab:sessions}, we see that there are only two sessions available for this subject, but the main reason for the models underperforming is that this subject's data have mainly PSPI intensity 1, which is extremely difficult to distinguish from no pain (PSPI=0). We further inspected the pain estimates for this subject, and all the models mainly produced the "no pain" label, which negatively affected the computation of the F-1 score for this subject. We also note that for subject \#59 all of the models produced the same estimates as this subject has only one session of data; thus, no adaptation was possible. In other words, when testing on session 1, all of the models' parameters are identical, since they were just initialized with the centrally pre-trained model parameters, thus yielding identical test scores for all models for this session. Also, for subject \#80, there is a drop in performance by all of the adapted models compared to the pre-trained BCDL model. We attribute this to the fact that, even though this subject has multiple sessions and active pain levels, the levels are mainly low PSPI levels of pain, the majority of which were PSPI=1. This, again, makes it difficult to separate the face images of this subject from no-pain images. 

Overall, we note that the traditional FL (FDL) and centralized learning (CDL) achieve similar performance. This is promising and it is due to the fact that FDL has flexibility to adjust the model parameters to each subject prior to the FL averaging, without seeing all raw image data. Furthermore, the locally trained models perform slightly worse on average, but when there are improvements, they can be large (e.g., see subject \#43). However, the performance of LDL is bounded by the amount of individual data, and has no means of recovering from severe overfitting that may occur (as in the case of subjects \#120 and \#80). We observe a similar trend in average performance across different metrics reported in Table~\ref{tab:scores-person}. Note that even-though the accuracy is the highest for the PFDL (by a small margin), this metric is less robust to unbalanced data than the F-1 score as it does not provide insights into how well the models are estimating (positive) pain - high accuracy can be achieved even if all data were estimated as no-pain due to this class largely dominating the dataset. On the other hand, precision-recall AUC is perhaps a more robust performance measure for unbalanced data. In this context, FDL outperforms the other models. However, to use this in practice would require finding optimal classification thresholds for each subject, which is difficult.  

\begin{table}[ht]
    \centering
    \begin{tabular}{lccc}
    \toprule
     \textbf{Method} &                \textbf{ACC} &   \textbf{PR-AUC} &       \textbf{F1} \\
     \midrule
         RND &            44 $\pm$ 15 &  31 $\pm$ 16 &   30 $\pm$ 13 \\
         BCDL &            72 $\pm$ 12 &  55 $\pm$ 16 &  40 $\pm$ 24 \\ \hline
         CDL &            73 $\pm$ 13 &  58 $\pm$ 16 &  46 $\pm$ 18 \\
         LDL &          75 $\pm$ 13 &  56 $\pm$ 18 &  44 $\pm$ 19 \\
          FDL &            73 $\pm$ 12 &  \textbf{59 $\pm$ 17} &  46 $\pm$ 20 \\ \hline
        PFDL &   \textbf{76 $\pm$ 12} &  57 $\pm$ 18 &  \textbf{47 $\pm$ 20}\\   
    \bottomrule
    \end{tabular}
        \caption{Average ($\pm$std) accuracy (ACC), precision-recall area under the curve (PR-AUC), and F1 score, across different sessions/test subjects. The high std reflects large variability in the performance per test subject.}
    \label{tab:scores-person}
\end{table}
It is important to emphasize that even though the average performance of these models is low, the models are evaluated on highly unbalanced data. Also, we used the raw face images without expert-driven processing (e.g., without using facial landmarks or face alignment). We did that to reduce computational cost, but we expect that additional image processing would significantly increase the estimation performance. However, the focus here was on assessing the benefits of different FL strategies, especially when the personalizing layer in the model is not shared, but is kept only locally.

\begin{figure}
    \centering
    \includegraphics[width=.35\textwidth]{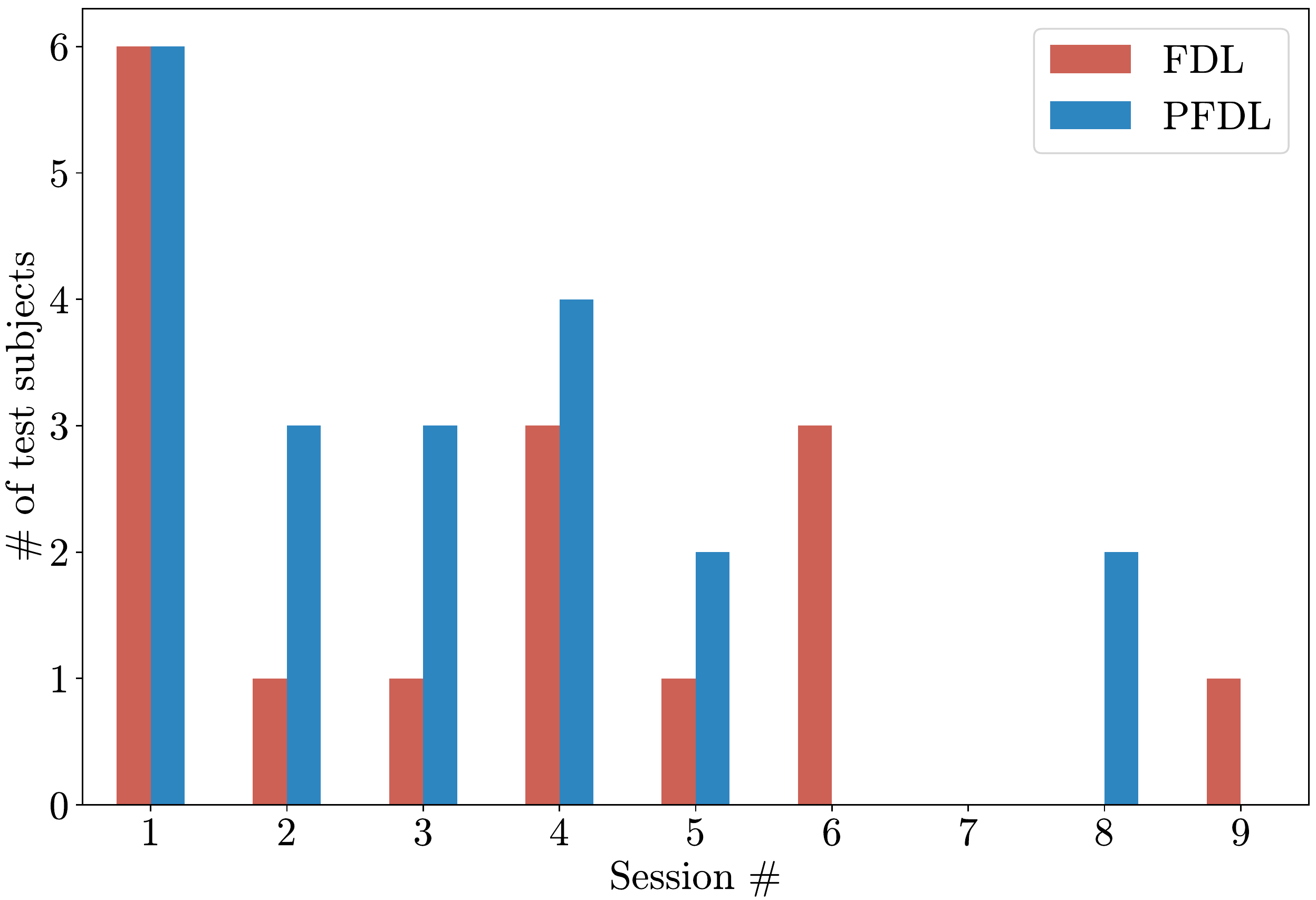}
    \caption{Personalized vs. standard federated learning  of deep models for pain estimation. The bars show on how many subjects one model outperformed the other (in terms of F-1) in each session.}
    \label{fig:federated-sessions}
\end{figure}

We next focus on F-1 performance of the FDL and PFDL models per session. In Fig.~\ref{fig:federated-sessions}, we show on how many test subjects one model outperformed the other per session. Both models perform the same on the first session as there is still no adaptation of the models. We then note that in sessions 2-5 the PFDL outperforms the standard FDL, while in sessions 8 and 9 each of the models outperforms the other. 
Drawing conclusions from these two sessions would be misleading as there are only one or two subjects available in these sessions. (Note session 7 had no test subject data).   

Finally, we analyze behaviour of the models during training by inspecting their training /validation loss during adaptation across different sessions. In all the models, the validation loss spikes substantially in session 5. In session 5, the number of examples with PSPI=1 is much higher than for the other levels (PSPI=2-15). Again, due to the very small changes in facial expression due to the pain level "1", it is very hard to separate it from pain level "0" (no pain). The validation loss also drops close to 0 in session 7. Looking at Table~\ref{tab:sessions}, we find that there are no positive examples in session 7. Consequently, the model becomes good at identifying true negative examples (no pain). Note that this is much more pronounced in PFDL and LDL models, which, in contrast to CDL and FDL, perform local fine-tuning of their last layer, most likely overfitting the adaptation set. As mentioned above, sessions 8 and 9 cannot provide reliable insights, as they contain only one or two participating test subjects. However, we note that FDL manages to reduce the validation loss, while PFDL and LDL keep increasing the validation loss during their adaptation to the one subject (\#52) in session 9. We suspect that this is a consequence of overfitting, which occurs in these two models while trying to train the last (classification) layer on the data of the previous session of that subject (using only a small number of positive examples from session 6 of this subject).

\begin{figure*}
    \centering
    \includegraphics[width=0.7\textwidth]{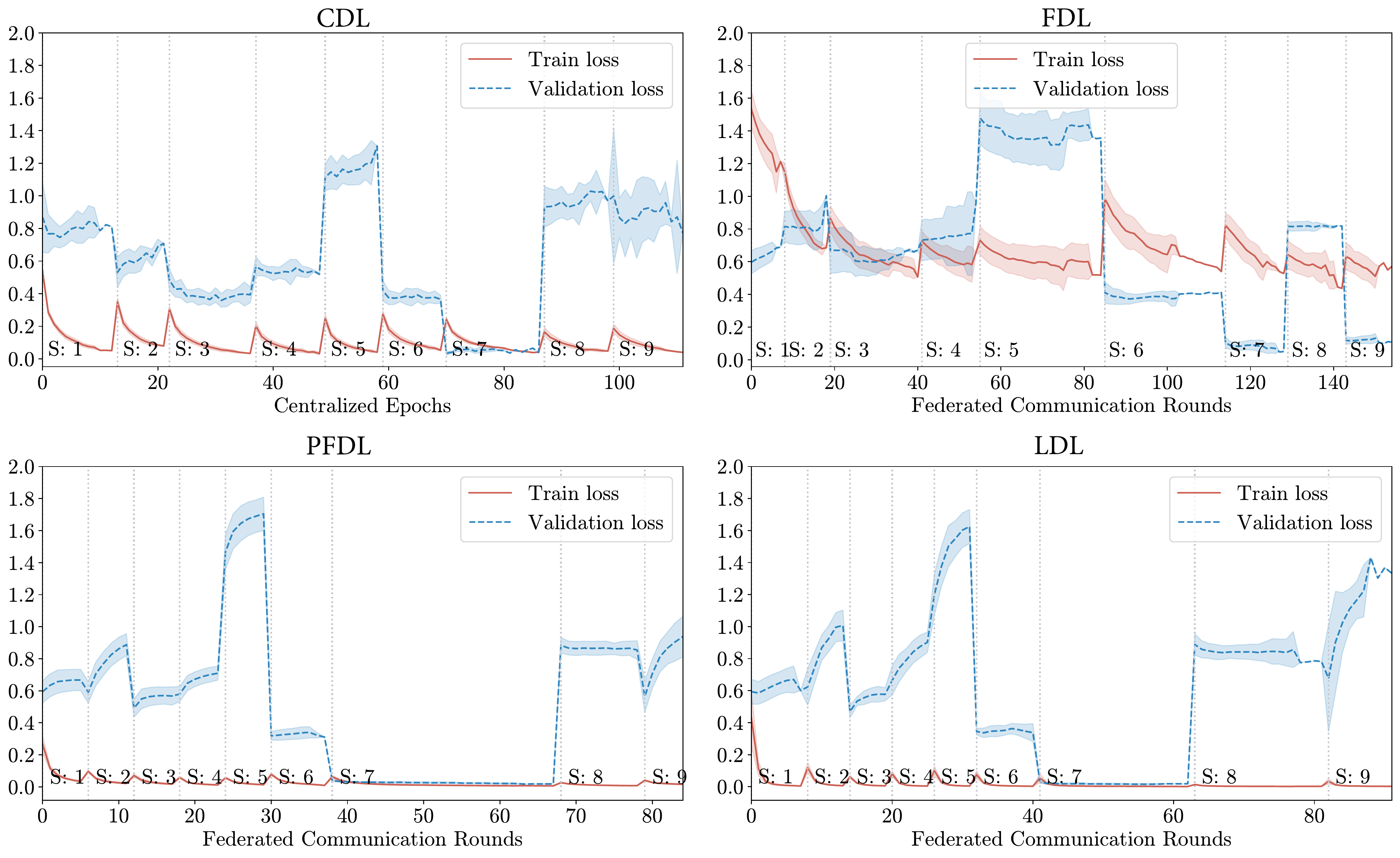}
    \caption{Average ($\pm$ 1 std) training/validation loss of ten different runs of the models, during communication rounds with different recording sessions.}
    \label{fig:train-val-loss}
\end{figure*}

\section{Discussion}
In real-world applications, the pain rating can be provided by clinical staff or via self-reports (e.g., using the traditional Visual Analog Scale) for a limited number of short video clips of pain expressions of the target subject. Using machine learning, and federated learning, as shown in this work, could enable the system to adapt to the target subject over time and thus reduce the need for manual labeling of the pain face-images. However, pain estimation from face images is difficult since the pain classifier relies on very subtle changes in facial expressions of target subjects. We showed that a light-weight CNN that adapts its parameters to test subjects, can largely outperform a pre-trained model that does not leverage the target subject's data. While this is expected, we showed it empirically on a challenging real-world pain dataset. More importantly, we showed that FL can achieve robust estimation of pain that is comparable to that achieved by the vanilla centralized training of the models. FL also tends to produce a more stable learning curve (see Fig.~\ref{fig:federated-sessions}), most likely because the model averaging has a similar effect as common regularization techniques such as dropout. Our findings on whether standard FL can yield substantially better results compared to centralized learning for a majority of clients are still not definite as the data contain a relatively small number of subjects. However, based on the results, we conjecture that in a federated setting, the data of subjects showing facial expressions of more intense pain experience nudge the average model more strongly in their favour.

Interestingly, the subject-specific models that did not participate in FL, still achieved comparable average results in terms of F-1 score, and better than the non-adapted pre-trained models. These results support the value of individual data and the potential of model personalization. In the experiments, we saw that these models largely outperformed the FL and centralized-learning models on some subjects, despite the fact that they were trained using only a handful of adaptation data of the target subject. While this may seem a promising approach (to use only the subject-specific models), these models do not have means for recovering from corrupt data (e.g., incorrect and/or highly noisy labels), as they rely fully on the local data. Conversely, the personalized FL approach proposed here combines the best of the two approaches: the personalization of the local models and the resilience of the models trained using the standard FL. 
Our experiments revealed that this \textit{federated personalization} rendered models that were not always performing the best. However, they consistently followed the best performing model, even in cases where other models would largely underperform. This is promising as it shows that by using \textit{federated personalization} we can obtain overall more robust pain estimation, while also providing another layer of data confidentiality (by not sharing the pain estimation layer). 

Regarding data-privacy and -confidentiality concerns, of course we cannot make any guarantees, as there could always be a breach locally. However, with respect to the shared data, our approach adds another confidentiality layer by not sharing as much as FL usually shares. Specifically, only the CNN layers in the network participate in FL but not the last fully-connected layer used to make a decision about the presence/absence of pain. Since this boils down to exchanging the parameters of the CNN kernels, and not their feature maps (i.e., the network output activations) or fully connected layers, it would be extremely difficult\footnote{By using the shared CNN parameters {\it only} (i.e., the CNN weights) it would not be possible to re-construct the original image of the target person. To our knowledge, no prior work has attempted this.}
for an adversary to reconstruct the face images of the target subject, let alone infer their pain labels. This, however, does not mean that in this setting the subject's ID (i.e., where the data is coming from, not necessarily from whom) cannot be identified - the shared parameters may still contain individual biases that could be inferred by an adversary. Still, the adversary cannot see any raw data of the target subject without breaching the subject's local storage. In cases where inferring the subject's ID is a breach of privacy (as it would be in most cases), methods based on differential privacy~\cite{dwork2014algorithmic} can be used to further test and secure the privacy-preserving properties of our approach. We plan to investigate this in our future work.

Another important aspect of our analysis concerns the pain dataset used in this work. This dataset poses a number of challenges that may have obscured some insights into the models that we investigated - as we describe in what follows. However, we wanted to evaluate the robustness of FL on publicly available real-world data. The difference between the face images of lower levels of pain and where test subjects do not experience any pain may be very nuanced. This, it turn, makes it challenging to estimate pain from such images even when the centralized training is applied. Moreover, the dataset is largely imbalanced and sparse in terms of the pain vs. no pain images, as shown in Table~\ref{tab:sessions}. To improve the baseline model of the learning algorithms under study, models that encode temporal dependencies in data (e.g.,  LSTMs), could be used in the FL settings. This and other more advanced base models have potential to better discriminate low levels of pain from no pain by having access to temporal context, which is one of the limitations of the currently used CNN architecture. To address the strong imbalance of the pain labels in the dataset, implementing different loss functions such as \textit{hinge loss}, or \textit{weighted binary cross-entropy}, may yield performance improvements. Yet, we believe that having access to richer pain datasets -- in terms of the number of subjects and their recording sessions over a longer period of time, as well as the pain ratings for the face-images/videos, is critical for achieving pain estimation performance that is needed for real-world applications. To our knowledge, such data are not readily available, and are challenging to collect and annotate. We expect that this will change in the near future as the mobile and wearable technologies become more pervasive \cite{kumar2020trustworthy}. Nevertheless, note that the focus of this paper was not to outperform the state-of-the-art systems for pain estimation; rather, the aim of this work was to benchmark centralized and federated learning algorithms on the challenging problem of pain estimation from face images, and identify some of the key challenges in tackling this problem using FL.  

There is certainly room for improvement of the federated personalization algorithm proposed in this work. In some sessions, there were subjects for whom the models experienced a decrease in performance compared to the pre-trained baseline model. This decrease is likely to be caused by some of the new sessions' data that the models were trained on (most likely, overfitted to), resulting in {\it negative transfer}~\cite{torrey2010transfer}. This typically occurs when the original data distribution is very different from the target data distribution. While we hoped that by adapting the model using data of previous sessions of the target subject the performance would constantly increase, in practice this is not the case. Even the data (the pain expressions) of the same subject may vary largely from session to session, thus, hurting the model performance after adaptation to such data. To prevent or reduce such negative transfer, more sophisticated model validation strategies need to be designed. Also, implementing a "fallback"-model in a federated setting that would prevent the updated model from underperforming on the future sessions, could reduce the negative transfer that our approach is currently sensitive to. Finally, evaluating our approach on health datasets that contain considerably more subjects would hopefully reveal the full potential of this approach. These challenges are the focus of our ongoing research.
\section{Conclusions}
We presented an evolution of the federated averaging algorithm, which we term \textit{federated personalization}. This algorithm adds one more layer of confidentiality to the traditional FL algorithm, by allowing only a fixed subset of the model parameters to be shared with a centralized server. We evaluated the new model by simulating a real-world setting in which the data of target subjects arrive in sessions, as would be the case in hospital facilities with remote and unobtrusive monitoring of patients' pain experience. Currently, this approach only modestly outperforms the traditional FL on a challenging problem of pain estimation from facial expressions; however, we discuss ways in which this approach can further be enhanced. The proposed is of high importance for healthcare domains, where regulations of confidentiality and privacy, as well as governance and accountability for individual data are top priorities.

\section*{Acknowledgement}
The work by O.Rudovic was conducted while funded by the European Commission Horizon 2020 programme, Marie Curie Action - Individual Fellowship 2016-2019 (EngageMe), under the agreement number 701236. The authors declare no competing interests.
\balance
\bibliographystyle{ACM-Reference-Format}
\bibliography{sample-base}


\begin{thebibliography}{49}


\ifx \showCODEN    \undefined \def \showCODEN     #1{\unskip}     \fi
\ifx \showDOI      \undefined \def \showDOI       #1{#1}\fi
\ifx \showISBNx    \undefined \def \showISBNx     #1{\unskip}     \fi
\ifx \showISBNxiii \undefined \def \showISBNxiii  #1{\unskip}     \fi
\ifx \showISSN     \undefined \def \showISSN      #1{\unskip}     \fi
\ifx \showLCCN     \undefined \def \showLCCN      #1{\unskip}     \fi
\ifx \shownote     \undefined \def \shownote      #1{#1}          \fi
\ifx \showarticletitle \undefined \def \showarticletitle #1{#1}   \fi
\ifx \showURL      \undefined \def \showURL       {\relax}        \fi
\providecommand\bibfield[2]{#2}
\providecommand\bibinfo[2]{#2}
\providecommand\natexlab[1]{#1}
\providecommand\showeprint[2][]{arXiv:#2}

\bibitem[\protect\citeauthoryear{??}{gdp}{[n.d.]}]%
        {gdpr}
 \bibinfo{year}{[n.d.]}\natexlab{}.
\newblock \bibinfo{title}{Guide to the General Data Protection Regulation
  (GDPR),
  https://ico.org.uk/for-organisations/guide-to-data-protection/guide-to-the-general-data-protection-regulation-gdpr/}.
\newblock
\newblock


\bibitem[\protect\citeauthoryear{??}{eco}{2019}]%
        {economist-nursing}
 \bibinfo{year}{2019}\natexlab{}.
\newblock \bibinfo{title}{A shortage of staff is the biggest problem facing the
  NHS,
  https://www.economist.com/britain/2019/03/23/a-shortage-of-staff-is-the-biggest-problem-facing-the-nhs}.
\newblock
\newblock


\bibitem[\protect\citeauthoryear{Ashraf, Lucey, Cohn, Chen, Ambadar, Prkachin,
  and Solomon}{Ashraf et~al\mbox{.}}{2009}]%
        {ashraf2009painful}
\bibfield{author}{\bibinfo{person}{Ahmed~Bilal Ashraf}, \bibinfo{person}{Simon
  Lucey}, \bibinfo{person}{Jeffrey~F Cohn}, \bibinfo{person}{Tsuhan Chen},
  \bibinfo{person}{Zara Ambadar}, \bibinfo{person}{Kenneth~M Prkachin}, {and}
  \bibinfo{person}{Patricia~E Solomon}.} \bibinfo{year}{2009}\natexlab{}.
\newblock \showarticletitle{The painful face--pain expression recognition using
  active appearance models}.
\newblock \bibinfo{journal}{\emph{Image and vision computing}}
  \bibinfo{volume}{27}, \bibinfo{number}{12} (\bibinfo{year}{2009}),
  \bibinfo{pages}{1788--1796}.
\newblock


\bibitem[\protect\citeauthoryear{Aung, Kaltwang, Romera-Paredes,
  et~al\mbox{.}}{Aung et~al\mbox{.}}{2015}]%
        {aung2015automatic}
\bibfield{author}{\bibinfo{person}{Min~SH Aung}, \bibinfo{person}{Sebastian
  Kaltwang}, \bibinfo{person}{Romera-Paredes}, {et~al\mbox{.}}}
  \bibinfo{year}{2015}\natexlab{}.
\newblock \showarticletitle{The automatic detection of chronic pain-related
  expression: requirements, challenges and the multimodal EmoPain dataset}.
\newblock \bibinfo{journal}{\emph{IEEE transactions on affective computing}}
  \bibinfo{volume}{7}, \bibinfo{number}{4} (\bibinfo{year}{2015}),
  \bibinfo{pages}{435--451}.
\newblock


\bibitem[\protect\citeauthoryear{Bonawitz, Eichner, Grieskamp, Huba, Ingerman,
  Ivanov, Kiddon, Konecny, Mazzocchi, McMahan, et~al\mbox{.}}{Bonawitz
  et~al\mbox{.}}{2019}]%
        {bonawitz2019towards}
\bibfield{author}{\bibinfo{person}{Keith Bonawitz}, \bibinfo{person}{Hubert
  Eichner}, \bibinfo{person}{Wolfgang Grieskamp}, \bibinfo{person}{Dzmitry
  Huba}, \bibinfo{person}{Alex Ingerman}, \bibinfo{person}{Vladimir Ivanov},
  \bibinfo{person}{Chloe Kiddon}, \bibinfo{person}{Jakub Konecny},
  \bibinfo{person}{Stefano Mazzocchi}, \bibinfo{person}{H~Brendan McMahan},
  {et~al\mbox{.}}} \bibinfo{year}{2019}\natexlab{}.
\newblock \showarticletitle{Towards federated learning at scale: System
  design}.
\newblock \bibinfo{journal}{\emph{arXiv preprint arXiv:1902.01046}}
  (\bibinfo{year}{2019}).
\newblock


\bibitem[\protect\citeauthoryear{Brisimi, Chen, Mela, Olshevsky, Paschalidis,
  and Shi}{Brisimi et~al\mbox{.}}{2018}]%
        {brisimi2018federated}
\bibfield{author}{\bibinfo{person}{Theodora~S Brisimi}, \bibinfo{person}{Ruidi
  Chen}, \bibinfo{person}{Theofanie Mela}, \bibinfo{person}{Alex Olshevsky},
  \bibinfo{person}{Ioannis~Ch Paschalidis}, {and} \bibinfo{person}{Wei Shi}.}
  \bibinfo{year}{2018}\natexlab{}.
\newblock \showarticletitle{Federated learning of predictive models from
  federated electronic health records}.
\newblock \bibinfo{journal}{\emph{International journal of medical
  informatics}}  \bibinfo{volume}{112} (\bibinfo{year}{2018}),
  \bibinfo{pages}{59--67}.
\newblock


\bibitem[\protect\citeauthoryear{Cervero}{Cervero}{2012}]%
        {cervero2012understanding}
\bibfield{author}{\bibinfo{person}{Fernando Cervero}.}
  \bibinfo{year}{2012}\natexlab{}.
\newblock \bibinfo{booktitle}{\emph{Understanding pain: exploring the
  perception of pain}}.
\newblock \bibinfo{publisher}{Mit Press}.
\newblock


\bibitem[\protect\citeauthoryear{Chen, Ansari, and Wilkie}{Chen
  et~al\mbox{.}}{2018}]%
        {chen2018automated}
\bibfield{author}{\bibinfo{person}{Zhanli Chen}, \bibinfo{person}{Rashid
  Ansari}, {and} \bibinfo{person}{Diana Wilkie}.}
  \bibinfo{year}{2018}\natexlab{}.
\newblock \showarticletitle{Automated Pain Detection from Facial Expressions
  using FACS: A Review}.
\newblock \bibinfo{journal}{\emph{arXiv preprint arXiv:1811.07988}}
  (\bibinfo{year}{2018}).
\newblock


\bibitem[\protect\citeauthoryear{Choudhury, Gkoulalas-Divanis, Salonidis,
  Sylla, Park, Hsu, and Das}{Choudhury et~al\mbox{.}}{2019a}]%
        {choudhury2019differential}
\bibfield{author}{\bibinfo{person}{Olivia Choudhury}, \bibinfo{person}{Aris
  Gkoulalas-Divanis}, \bibinfo{person}{Theodoros Salonidis},
  \bibinfo{person}{Issa Sylla}, \bibinfo{person}{Yoonyoung Park},
  \bibinfo{person}{Grace Hsu}, {and} \bibinfo{person}{Amar Das}.}
  \bibinfo{year}{2019}\natexlab{a}.
\newblock \showarticletitle{Differential privacy-enabled federated learning for
  sensitive health data}.
\newblock \bibinfo{journal}{\emph{arXiv preprint arXiv:1910.02578}}
  (\bibinfo{year}{2019}).
\newblock


\bibitem[\protect\citeauthoryear{Choudhury, Park, Salonidis, Gkoulalas-Divanis,
  Sylla, et~al\mbox{.}}{Choudhury et~al\mbox{.}}{2019b}]%
        {choudhury2019predicting}
\bibfield{author}{\bibinfo{person}{Olivia Choudhury},
  \bibinfo{person}{Yoonyoung Park}, \bibinfo{person}{Theodoros Salonidis},
  \bibinfo{person}{Aris Gkoulalas-Divanis}, \bibinfo{person}{Issa Sylla},
  {et~al\mbox{.}}} \bibinfo{year}{2019}\natexlab{b}.
\newblock \showarticletitle{Predicting Adverse Drug Reactions on Distributed
  Health Data using Federated Learning}.
\newblock   \bibinfo{volume}{2019} (\bibinfo{year}{2019}),
  \bibinfo{pages}{313}.
\newblock


\bibitem[\protect\citeauthoryear{Dwork, Roth, et~al\mbox{.}}{Dwork
  et~al\mbox{.}}{2014}]%
        {dwork2014algorithmic}
\bibfield{author}{\bibinfo{person}{Cynthia Dwork}, \bibinfo{person}{Aaron
  Roth}, {et~al\mbox{.}}} \bibinfo{year}{2014}\natexlab{}.
\newblock \showarticletitle{The algorithmic foundations of differential
  privacy}.
\newblock \bibinfo{journal}{\emph{Foundations and Trends{\textregistered} in
  Theoretical Computer Science}} \bibinfo{volume}{9}, \bibinfo{number}{3--4}
  (\bibinfo{year}{2014}), \bibinfo{pages}{211--407}.
\newblock


\bibitem[\protect\citeauthoryear{Egede, Valstar, and Martinez}{Egede
  et~al\mbox{.}}{2017}]%
        {egede2017fusing}
\bibfield{author}{\bibinfo{person}{Joy Egede}, \bibinfo{person}{Michel
  Valstar}, {and} \bibinfo{person}{Brais Martinez}.}
  \bibinfo{year}{2017}\natexlab{}.
\newblock \showarticletitle{Fusing deep learned and hand-crafted features of
  appearance, shape, and dynamics for automatic pain estimation}. In
  \bibinfo{booktitle}{\emph{2017 12th IEEE international conference on
  automatic face \& gesture recognition (FG 2017)}}. IEEE,
  \bibinfo{pages}{689--696}.
\newblock


\bibitem[\protect\citeauthoryear{Ekman, Friesen, and Hager}{Ekman
  et~al\mbox{.}}{2002}]%
        {ekman2002facial}
\bibfield{author}{\bibinfo{person}{P Ekman}, \bibinfo{person}{W Friesen}, {and}
  \bibinfo{person}{J Hager}.} \bibinfo{year}{2002}\natexlab{}.
\newblock \showarticletitle{Facial action coding system: Research Nexus}.
\newblock \bibinfo{journal}{\emph{Network Research Information, Salt Lake City,
  UT}}  \bibinfo{volume}{1} (\bibinfo{year}{2002}).
\newblock


\bibitem[\protect\citeauthoryear{Gawande}{Gawande}{2010}]%
        {checklist-manifesto}
\bibfield{author}{\bibinfo{person}{Atul Gawande}.}
  \bibinfo{year}{2010}\natexlab{}.
\newblock \bibinfo{booktitle}{\emph{Checklist manifesto, the (HB)}}.
\newblock \bibinfo{publisher}{Penguin Books India}.
\newblock


\bibitem[\protect\citeauthoryear{Goodfellow, Pouget-Abadie, Mirza, Xu,
  Warde-Farley, Ozair, Courville, and Bengio}{Goodfellow et~al\mbox{.}}{2014}]%
        {goodfellow2014generative}
\bibfield{author}{\bibinfo{person}{Ian Goodfellow}, \bibinfo{person}{Jean
  Pouget-Abadie}, \bibinfo{person}{Mehdi Mirza}, \bibinfo{person}{Bing Xu},
  \bibinfo{person}{David Warde-Farley}, \bibinfo{person}{Sherjil Ozair},
  \bibinfo{person}{Aaron Courville}, {and} \bibinfo{person}{Yoshua Bengio}.}
  \bibinfo{year}{2014}\natexlab{}.
\newblock \showarticletitle{Generative adversarial nets}. In
  \bibinfo{booktitle}{\emph{Advances in neural information processing
  systems}}. \bibinfo{pages}{2672--2680}.
\newblock


\bibitem[\protect\citeauthoryear{Hammal and Cohn}{Hammal and Cohn}{2018}]%
        {hammal2018automatic}
\bibfield{author}{\bibinfo{person}{Zakia Hammal} {and}
  \bibinfo{person}{Jeffrey~F Cohn}.} \bibinfo{year}{2018}\natexlab{}.
\newblock \showarticletitle{Automatic, objective, and efficient measurement of
  pain using automated face analysis}.
\newblock In \bibinfo{booktitle}{\emph{Social and Interpersonal Dynamics in
  Pain}}. \bibinfo{publisher}{Springer}, \bibinfo{pages}{121--146}.
\newblock


\bibitem[\protect\citeauthoryear{Hammal and Kunz}{Hammal and Kunz}{2012}]%
        {hammal2012pain}
\bibfield{author}{\bibinfo{person}{Zakia Hammal} {and} \bibinfo{person}{Miriam
  Kunz}.} \bibinfo{year}{2012}\natexlab{}.
\newblock \showarticletitle{Pain monitoring: A dynamic and context-sensitive
  system}.
\newblock \bibinfo{journal}{\emph{Pattern Recognition}} \bibinfo{volume}{45},
  \bibinfo{number}{4} (\bibinfo{year}{2012}), \bibinfo{pages}{1265--1280}.
\newblock


\bibitem[\protect\citeauthoryear{Hardy, Henecka, Ivey{-}Law, Nock, Patrini,
  Smith, and Thorne}{Hardy et~al\mbox{.}}{2017}]%
        {federated-log-regression}
\bibfield{author}{\bibinfo{person}{Stephen Hardy}, \bibinfo{person}{Wilko
  Henecka}, \bibinfo{person}{Hamish Ivey{-}Law}, \bibinfo{person}{Richard
  Nock}, \bibinfo{person}{Giorgio Patrini}, \bibinfo{person}{Guillaume Smith},
  {and} \bibinfo{person}{Brian Thorne}.} \bibinfo{year}{2017}\natexlab{}.
\newblock \showarticletitle{Private federated learning on vertically
  partitioned data via entity resolution and additively homomorphic
  encryption}.
\newblock \bibinfo{journal}{\emph{CoRR}}  \bibinfo{volume}{abs / 1711.10677}
  (\bibinfo{year}{2017}).
\newblock
\showeprint[arxiv]{1711.10677}
\urldef\tempurl%
\url{http://arxiv.org/abs/1711.10677}
\showURL{%
\tempurl}


\bibitem[\protect\citeauthoryear{Hartmann}{Hartmann}{2018}]%
        {hartmann2018federated}
\bibfield{author}{\bibinfo{person}{Florian Hartmann}.}
  \bibinfo{year}{2018}\natexlab{}.
\newblock \bibinfo{title}{Federated Learning}.
\newblock
\newblock


\bibitem[\protect\citeauthoryear{He, Zhang, Ren, and Sun}{He
  et~al\mbox{.}}{2015}]%
        {resnet}
\bibfield{author}{\bibinfo{person}{Kaiming He}, \bibinfo{person}{Xiangyu
  Zhang}, \bibinfo{person}{Shaoqing Ren}, {and} \bibinfo{person}{Jian Sun}.}
  \bibinfo{year}{2015}\natexlab{}.
\newblock \showarticletitle{Deep Residual Learning for Image Recognition}.
\newblock \bibinfo{journal}{\emph{CoRR}}  \bibinfo{volume}{abs/1512.03385}
  (\bibinfo{year}{2015}).
\newblock
\showeprint[arxiv]{1512.03385}
\urldef\tempurl%
\url{http://arxiv.org/abs/1512.03385}
\showURL{%
\tempurl}


\bibitem[\protect\citeauthoryear{HHS Office of~the Secretary and Ocr}{HHS
  Office of~the Secretary and Ocr}{2013}]%
        {hipaa}
\bibfield{author}{\bibinfo{person}{Office for Civil~Rights HHS Office of~the
  Secretary} {and} \bibinfo{person}{Ocr}.} \bibinfo{year}{2013}\natexlab{}.
\newblock \bibinfo{title}{Summary of the HIPAA Security Rule, HHS.gov, The US
  Department of Health and Human Services,
  https://www.hhs.gov/hipaa/for-professionals/security/laws-regulations/index.html}.
\newblock
\newblock


\bibitem[\protect\citeauthoryear{Hitaj, Ateniese, and Perez-Cruz}{Hitaj
  et~al\mbox{.}}{2017}]%
        {hitaj2017deep}
\bibfield{author}{\bibinfo{person}{Briland Hitaj}, \bibinfo{person}{Giuseppe
  Ateniese}, {and} \bibinfo{person}{Fernando Perez-Cruz}.}
  \bibinfo{year}{2017}\natexlab{}.
\newblock \showarticletitle{Deep models under the GAN: information leakage from
  collaborative deep learning}. In \bibinfo{booktitle}{\emph{Proceedings of the
  2017 ACM SIGSAC Conference on Computer and Communications Security}}. ACM,
  \bibinfo{pages}{603--618}.
\newblock


\bibitem[\protect\citeauthoryear{Howard, Zhu, Chen, Kalenichenko, Wang, Weyand,
  Andreetto, and Adam}{Howard et~al\mbox{.}}{2017}]%
        {mobilenets}
\bibfield{author}{\bibinfo{person}{Andrew~G. Howard}, \bibinfo{person}{Menglong
  Zhu}, \bibinfo{person}{Bo Chen}, \bibinfo{person}{Dmitry Kalenichenko},
  \bibinfo{person}{Weijun Wang}, \bibinfo{person}{Tobias Weyand},
  \bibinfo{person}{Marco Andreetto}, {and} \bibinfo{person}{Hartwig Adam}.}
  \bibinfo{year}{2017}\natexlab{}.
\newblock \showarticletitle{MobileNets: Efficient Convolutional Neural Networks
  for Mobile Vision Applications}.
\newblock \bibinfo{journal}{\emph{CoRR}}  \bibinfo{volume}{abs/1704.04861}
  (\bibinfo{year}{2017}).
\newblock
\showeprint[arxiv]{1704.04861}
\urldef\tempurl%
\url{http://arxiv.org/abs/1704.04861}
\showURL{%
\tempurl}


\bibitem[\protect\citeauthoryear{Kairouz, McMahan, Avent, Bellet, Bennis,
  Bhagoji, Bonawitz, Charles, Cormode, Cummings, et~al\mbox{.}}{Kairouz
  et~al\mbox{.}}{2019}]%
        {kairouz2019advances}
\bibfield{author}{\bibinfo{person}{Peter Kairouz}, \bibinfo{person}{H~Brendan
  McMahan}, \bibinfo{person}{Brendan Avent}, \bibinfo{person}{Aur{\'e}lien
  Bellet}, \bibinfo{person}{Mehdi Bennis}, \bibinfo{person}{Arjun~Nitin
  Bhagoji}, \bibinfo{person}{KA Bonawitz}, \bibinfo{person}{Zachary Charles},
  \bibinfo{person}{Graham Cormode}, \bibinfo{person}{Rachel Cummings},
  {et~al\mbox{.}}} \bibinfo{year}{2019}\natexlab{}.
\newblock \showarticletitle{Advances and Open Problems in Federated Learning}.
\newblock  (\bibinfo{year}{2019}).
\newblock


\bibitem[\protect\citeauthoryear{Kaltwang, Rudovic, and Pantic}{Kaltwang
  et~al\mbox{.}}{2012}]%
        {kaltwang2012continuous}
\bibfield{author}{\bibinfo{person}{Sebastian Kaltwang}, \bibinfo{person}{Ognjen
  Rudovic}, {and} \bibinfo{person}{Maja Pantic}.}
  \bibinfo{year}{2012}\natexlab{}.
\newblock \showarticletitle{Continuous pain intensity estimation from facial
  expressions}. In \bibinfo{booktitle}{\emph{International Symposium on Visual
  Computing}}. Springer, \bibinfo{pages}{368--377}.
\newblock


\bibitem[\protect\citeauthoryear{Kirkpatrick, Pascanu, Rabinowitz, Veness,
  Desjardins, Rusu, Milan, Quan, Ramalho, Grabska-Barwinska,
  et~al\mbox{.}}{Kirkpatrick et~al\mbox{.}}{2017}]%
        {kirkpatrick2017overcoming}
\bibfield{author}{\bibinfo{person}{James Kirkpatrick}, \bibinfo{person}{Razvan
  Pascanu}, \bibinfo{person}{Neil Rabinowitz}, \bibinfo{person}{Joel Veness},
  \bibinfo{person}{Guillaume Desjardins}, \bibinfo{person}{Andrei~A Rusu},
  \bibinfo{person}{Kieran Milan}, \bibinfo{person}{John Quan},
  \bibinfo{person}{Tiago Ramalho}, \bibinfo{person}{Agnieszka
  Grabska-Barwinska}, {et~al\mbox{.}}} \bibinfo{year}{2017}\natexlab{}.
\newblock \showarticletitle{Overcoming catastrophic forgetting in neural
  networks}.
\newblock \bibinfo{journal}{\emph{Proceedings of the national academy of
  sciences}} \bibinfo{volume}{114}, \bibinfo{number}{13}
  (\bibinfo{year}{2017}), \bibinfo{pages}{3521--3526}.
\newblock


\bibitem[\protect\citeauthoryear{Konecn{\'{y}}, McMahan, Ramage, and
  Richt{\'{a}}rik}{Konecn{\'{y}} et~al\mbox{.}}{2016}]%
        {ref-google-FML-init3}
\bibfield{author}{\bibinfo{person}{Jakub Konecn{\'{y}}},
  \bibinfo{person}{H.~Brendan McMahan}, \bibinfo{person}{Daniel Ramage}, {and}
  \bibinfo{person}{Peter Richt{\'{a}}rik}.} \bibinfo{year}{2016}\natexlab{}.
\newblock \showarticletitle{Federated Optimization: Distributed Machine
  Learning for On-Device Intelligence}.
\newblock \bibinfo{journal}{\emph{CoRR}}  \bibinfo{volume}{abs/1610.02527}
  (\bibinfo{year}{2016}).
\newblock
\showeprint[arxiv]{1610.02527}
\urldef\tempurl%
\url{http://arxiv.org/abs/1610.02527}
\showURL{%
\tempurl}


\bibitem[\protect\citeauthoryear{Kone{\v{c}}n{\`y}, McMahan, Yu, Richt{\'a}rik,
  Suresh, and Bacon}{Kone{\v{c}}n{\`y} et~al\mbox{.}}{2016}]%
        {ref-google-FML-init2}
\bibfield{author}{\bibinfo{person}{Jakub Kone{\v{c}}n{\`y}},
  \bibinfo{person}{H~Brendan McMahan}, \bibinfo{person}{Felix~X Yu},
  \bibinfo{person}{Peter Richt{\'a}rik}, \bibinfo{person}{Ananda~Theertha
  Suresh}, {and} \bibinfo{person}{Dave Bacon}.}
  \bibinfo{year}{2016}\natexlab{}.
\newblock \showarticletitle{Federated learning: Strategies for improving
  communication efficiency}.
\newblock \bibinfo{journal}{\emph{arXiv preprint arXiv:1610.05492}}
  (\bibinfo{year}{2016}).
\newblock


\bibitem[\protect\citeauthoryear{Kumar, Braud, Tarkoma, and Hui}{Kumar
  et~al\mbox{.}}{2020}]%
        {kumar2020trustworthy}
\bibfield{author}{\bibinfo{person}{Abhishek Kumar}, \bibinfo{person}{Tristan
  Braud}, \bibinfo{person}{Sasu Tarkoma}, {and} \bibinfo{person}{Pan Hui}.}
  \bibinfo{year}{2020}\natexlab{}.
\newblock \showarticletitle{Trustworthy AI in the Age of Pervasive Computing
  and Big Data}.
\newblock \bibinfo{journal}{\emph{arXiv preprint arXiv:2002.05657}}
  (\bibinfo{year}{2020}).
\newblock


\bibitem[\protect\citeauthoryear{Liu, Miller, Sayeed, and Mandl}{Liu
  et~al\mbox{.}}{2018}]%
        {liu2018fadl}
\bibfield{author}{\bibinfo{person}{Dianbo Liu}, \bibinfo{person}{Timothy
  Miller}, \bibinfo{person}{Raheel Sayeed}, {and} \bibinfo{person}{Kenneth~D
  Mandl}.} \bibinfo{year}{2018}\natexlab{}.
\newblock \showarticletitle{Fadl: Federated-autonomous deep learning for
  distributed electronic health record}.
\newblock \bibinfo{journal}{\emph{arXiv preprint arXiv:1811.11400}}
  (\bibinfo{year}{2018}).
\newblock


\bibitem[\protect\citeauthoryear{Lucey, Cohn, Matthews, Lucey, Sridharan,
  Howlett, and Prkachin}{Lucey et~al\mbox{.}}{2010}]%
        {lucey2010automatically}
\bibfield{author}{\bibinfo{person}{Patrick Lucey}, \bibinfo{person}{Jeffrey~F
  Cohn}, \bibinfo{person}{Iain Matthews}, \bibinfo{person}{Simon Lucey},
  \bibinfo{person}{Sridha Sridharan}, \bibinfo{person}{Jessica Howlett}, {and}
  \bibinfo{person}{Kenneth~M Prkachin}.} \bibinfo{year}{2010}\natexlab{}.
\newblock \showarticletitle{Automatically detecting pain in video through
  facial action units}.
\newblock \bibinfo{journal}{\emph{IEEE Transactions on Systems, Man, and
  Cybernetics, Part B (Cybernetics)}} \bibinfo{volume}{41}, \bibinfo{number}{3}
  (\bibinfo{year}{2010}), \bibinfo{pages}{664--674}.
\newblock


\bibitem[\protect\citeauthoryear{{Lucey}, {Cohn}, {Prkachin}, {Solomon}, and
  {Matthews}}{{Lucey} et~al\mbox{.}}{2011}]%
        {painful-data}
\bibfield{author}{\bibinfo{person}{P. {Lucey}}, \bibinfo{person}{J.~F. {Cohn}},
  \bibinfo{person}{K.~M. {Prkachin}}, \bibinfo{person}{P.~E. {Solomon}}, {and}
  \bibinfo{person}{I. {Matthews}}.} \bibinfo{year}{2011}\natexlab{}.
\newblock \showarticletitle{Painful data: The UNBC-McMaster shoulder pain
  expression archive database}. In \bibinfo{booktitle}{\emph{Face and Gesture
  2011}}. \bibinfo{pages}{57--64}.
\newblock
\urldef\tempurl%
\url{https://doi.org/10.1109/FG.2011.5771462}
\showDOI{\tempurl}


\bibitem[\protect\citeauthoryear{McMahan and Ramage}{McMahan and
  Ramage}{2017}]%
        {broadband}
\bibfield{author}{\bibinfo{person}{Brendan McMahan} {and}
  \bibinfo{person}{Daniel Ramage}.} \bibinfo{year}{2017}\natexlab{}.
\newblock \bibinfo{title}{Federated Learning: Collaborative Machine Learning
  without Centralized Training Data}.
\newblock
\newblock
\urldef\tempurl%
\url{http://ai.googleblog.com/2017/04/federated-learning-collaborative.html}
\showURL{%
\tempurl}


\bibitem[\protect\citeauthoryear{McMahan, Moore, Ramage, and y~Arcas}{McMahan
  et~al\mbox{.}}{2016}]%
        {ref-google-FML-init}
\bibfield{author}{\bibinfo{person}{H.~Brendan McMahan}, \bibinfo{person}{Eider
  Moore}, \bibinfo{person}{Daniel Ramage}, {and}
  \bibinfo{person}{Blaise~Ag{\"{u}}era y Arcas}.}
  \bibinfo{year}{2016}\natexlab{}.
\newblock \showarticletitle{Federated Learning of Deep Networks using Model
  Averaging}.
\newblock \bibinfo{journal}{\emph{CoRR}}  \bibinfo{volume}{abs/1602.05629}
  (\bibinfo{year}{2016}).
\newblock
\showeprint[arxiv]{1602.05629}
\urldef\tempurl%
\url{http://arxiv.org/abs/1602.05629}
\showURL{%
\tempurl}


\bibitem[\protect\citeauthoryear{Melis, Song, De~Cristofaro, and
  Shmatikov}{Melis et~al\mbox{.}}{2018}]%
        {melis2018exploiting}
\bibfield{author}{\bibinfo{person}{Luca Melis}, \bibinfo{person}{Congzheng
  Song}, \bibinfo{person}{Emiliano De~Cristofaro}, {and}
  \bibinfo{person}{Vitaly Shmatikov}.} \bibinfo{year}{2018}\natexlab{}.
\newblock \showarticletitle{Exploiting unintended feature leakage in
  collaborative learning}.
\newblock \bibinfo{journal}{\emph{arXiv preprint arXiv:1805.04049}}
  (\bibinfo{year}{2018}).
\newblock


\bibitem[\protect\citeauthoryear{Poirot, Vepakomma, Chang, Kalpathy-Cramer,
  Gupta, and Raskar}{Poirot et~al\mbox{.}}{2019}]%
        {poirot2019split}
\bibfield{author}{\bibinfo{person}{Maarten~G Poirot}, \bibinfo{person}{Praneeth
  Vepakomma}, \bibinfo{person}{Ken Chang}, \bibinfo{person}{Jayashree
  Kalpathy-Cramer}, \bibinfo{person}{Rajiv Gupta}, {and}
  \bibinfo{person}{Ramesh Raskar}.} \bibinfo{year}{2019}\natexlab{}.
\newblock \showarticletitle{Split Learning for collaborative deep learning in
  healthcare}.
\newblock \bibinfo{journal}{\emph{arXiv preprint arXiv:1912.12115}}
  (\bibinfo{year}{2019}).
\newblock


\bibitem[\protect\citeauthoryear{Prkachin}{Prkachin}{1992}]%
        {prkachin1992consistency}
\bibfield{author}{\bibinfo{person}{Kenneth~M Prkachin}.}
  \bibinfo{year}{1992}\natexlab{}.
\newblock \showarticletitle{The consistency of facial expressions of pain: a
  comparison across modalities}.
\newblock \bibinfo{journal}{\emph{Pain}} \bibinfo{volume}{51},
  \bibinfo{number}{3} (\bibinfo{year}{1992}), \bibinfo{pages}{297--306}.
\newblock


\bibitem[\protect\citeauthoryear{Prkachin and Solomon}{Prkachin and
  Solomon}{2008}]%
        {prkachin2008structure}
\bibfield{author}{\bibinfo{person}{Kenneth~M Prkachin} {and}
  \bibinfo{person}{Patricia~E Solomon}.} \bibinfo{year}{2008}\natexlab{}.
\newblock \showarticletitle{The structure, reliability and validity of pain
  expression: Evidence from patients with shoulder pain}.
\newblock \bibinfo{journal}{\emph{Pain}} \bibinfo{volume}{139},
  \bibinfo{number}{2} (\bibinfo{year}{2008}), \bibinfo{pages}{267--274}.
\newblock


\bibitem[\protect\citeauthoryear{Rieke, Hancox, Li, Milletari, Roth,
  Albarqouni, Bakas, Galtier, Landman, Maier-Hein, Ourselin, Sheller, Summers,
  Trask, Xu, Baust, and Cardoso}{Rieke et~al\mbox{.}}{2020}]%
        {Rieke2020TheFO}
\bibfield{author}{\bibinfo{person}{Nicola Rieke}, \bibinfo{person}{Jonny
  Hancox}, \bibinfo{person}{Wenqi Li}, \bibinfo{person}{F. Milletari},
  \bibinfo{person}{H. Roth}, \bibinfo{person}{Shadi Albarqouni},
  \bibinfo{person}{S. Bakas}, \bibinfo{person}{M. Galtier}, \bibinfo{person}{B.
  Landman}, \bibinfo{person}{Klaus Maier-Hein}, \bibinfo{person}{S{\'e}bastien
  Ourselin}, \bibinfo{person}{Micah~J. Sheller}, \bibinfo{person}{R. Summers},
  \bibinfo{person}{Andrew Trask}, \bibinfo{person}{Daguang Xu},
  \bibinfo{person}{Maximilian Baust}, {and} \bibinfo{person}{M. Cardoso}.}
  \bibinfo{year}{2020}\natexlab{}.
\newblock \showarticletitle{The future of digital health with federated
  learning}.
\newblock \bibinfo{journal}{\emph{NPJ Digital Medicine}}  \bibinfo{volume}{3}
  (\bibinfo{year}{2020}).
\newblock


\bibitem[\protect\citeauthoryear{Rodriguez, Cucurull, Gonz{\`a}lez, Gonfaus,
  Nasrollahi, Moeslund, and Roca}{Rodriguez et~al\mbox{.}}{2017}]%
        {rodriguez2017deep}
\bibfield{author}{\bibinfo{person}{Pau Rodriguez}, \bibinfo{person}{Guillem
  Cucurull}, \bibinfo{person}{Jordi Gonz{\`a}lez}, \bibinfo{person}{Josep~M
  Gonfaus}, \bibinfo{person}{Kamal Nasrollahi}, \bibinfo{person}{Thomas~B
  Moeslund}, {and} \bibinfo{person}{F~Xavier Roca}.}
  \bibinfo{year}{2017}\natexlab{}.
\newblock \showarticletitle{Deep pain: Exploiting long short-term memory
  networks for facial expression classification}.
\newblock \bibinfo{journal}{\emph{IEEE transactions on cybernetics}}
  (\bibinfo{year}{2017}).
\newblock


\bibitem[\protect\citeauthoryear{Sheller, Reina, Edwards, Martin, and
  Bakas}{Sheller et~al\mbox{.}}{2018}]%
        {fed-intel}
\bibfield{author}{\bibinfo{person}{Micah~J. Sheller},
  \bibinfo{person}{G.~Anthony Reina}, \bibinfo{person}{Brandon Edwards},
  \bibinfo{person}{Jason Martin}, {and} \bibinfo{person}{Spyridon Bakas}.}
  \bibinfo{year}{2018}\natexlab{}.
\newblock \showarticletitle{Multi-Institutional Deep Learning Modeling Without
  Sharing Patient Data: {A} Feasibility Study on Brain Tumor Segmentation}.
\newblock \bibinfo{journal}{\emph{CoRR}}  \bibinfo{volume}{abs/1810.04304}
  (\bibinfo{year}{2018}).
\newblock
\showeprint[arxiv]{1810.04304}
\urldef\tempurl%
\url{http://arxiv.org/abs/1810.04304}
\showURL{%
\tempurl}


\bibitem[\protect\citeauthoryear{Simonyan and Zisserman}{Simonyan and
  Zisserman}{2014}]%
        {vgg}
\bibfield{author}{\bibinfo{person}{Karen Simonyan} {and}
  \bibinfo{person}{Andrew Zisserman}.} \bibinfo{year}{2014}\natexlab{}.
\newblock \showarticletitle{Very deep convolutional networks for large-scale
  image recognition}.
\newblock \bibinfo{journal}{\emph{arXiv preprint arXiv:1409.1556}}
  (\bibinfo{year}{2014}).
\newblock


\bibitem[\protect\citeauthoryear{Soar, Bargshady, Zhou, and Whittaker}{Soar
  et~al\mbox{.}}{2018}]%
        {soar2018deep}
\bibfield{author}{\bibinfo{person}{Jeffrey Soar}, \bibinfo{person}{Ghazal
  Bargshady}, \bibinfo{person}{Xujuan Zhou}, {and} \bibinfo{person}{Frank
  Whittaker}.} \bibinfo{year}{2018}\natexlab{}.
\newblock \showarticletitle{Deep learning model for detection of pain intensity
  from facial expression}. In \bibinfo{booktitle}{\emph{International
  Conference on Smart Homes and Health Telematics}}. Springer,
  \bibinfo{pages}{249--254}.
\newblock


\bibitem[\protect\citeauthoryear{Torrey and Shavlik}{Torrey and
  Shavlik}{2010}]%
        {torrey2010transfer}
\bibfield{author}{\bibinfo{person}{Lisa Torrey} {and} \bibinfo{person}{Jude
  Shavlik}.} \bibinfo{year}{2010}\natexlab{}.
\newblock \showarticletitle{Transfer learning}.
\newblock In \bibinfo{booktitle}{\emph{Handbook of research on machine learning
  applications and trends: algorithms, methods, and techniques}}.
  \bibinfo{publisher}{IGI Global}, \bibinfo{pages}{242--264}.
\newblock


\bibitem[\protect\citeauthoryear{{Walecki}, {Ognjen}, {Rudovic}, {Pavlovic},
  {Schuller}, and {Pantic}}{{Walecki} et~al\mbox{.}}{[n.d.]}]%
        {facial-action-unit}
\bibfield{author}{\bibinfo{person}{Robert {Walecki}},
  \bibinfo{person}{{Ognjen}}, \bibinfo{person}{{Rudovic}},
  \bibinfo{person}{Vladimir {Pavlovic}}, \bibinfo{person}{Bj{\"o}rn
  {Schuller}}, {and} \bibinfo{person}{Maja {Pantic}}.}
  \bibinfo{year}{[n.d.]}\natexlab{}.
\newblock \showarticletitle{{Deep Structured Learning for Facial Action Unit
  Intensity Estimation}}.
\newblock \bibinfo{journal}{\emph{CVPR, 2017}} (\bibinfo{year}{[n.\,d.]}).
\newblock


\bibitem[\protect\citeauthoryear{Wei, Wang, Bradfield, Li, Cardinale,
  Frackelton, Kim, Mentch, Van Steen, Visscher, Baldassano, and
  Hakonarson}{Wei et~al\mbox{.}}{2013}]%
        {rare-disease}
\bibfield{author}{\bibinfo{person}{Zhi Wei}, \bibinfo{person}{Wei Wang},
  \bibinfo{person}{Jonathan Bradfield}, \bibinfo{person}{Jin Li},
  \bibinfo{person}{Christopher Cardinale}, \bibinfo{person}{Edward Frackelton},
  \bibinfo{person}{Cecilia Kim}, \bibinfo{person}{Frank Mentch},
  \bibinfo{person}{Kristel Van Steen}, \bibinfo{person}{Peter M. Visscher},
  \bibinfo{person}{Robert N. Baldassano}, {and} \bibinfo{person}{Hakon
  Hakonarson}.} \bibinfo{year}{2013}\natexlab{}.
\newblock \showarticletitle{Large Sample Size, Wide Variant Spectrum, and
  Advanced Machine-Learning Technique Boost Risk Prediction for Inflammatory
  Bowel Disease}.
\newblock \bibinfo{journal}{\emph{The American Journal of Human Genetics}}
  \bibinfo{volume}{92}, \bibinfo{number}{6} (\bibinfo{year}{2013}),
  \bibinfo{pages}{1008 -- 1012}.
\newblock
\showISSN{0002-9297}
\urldef\tempurl%
\url{https://doi.org/10.1016/j.ajhg.2013.05.002}
\showDOI{\tempurl}


\bibitem[\protect\citeauthoryear{Wells, Pasero, and Mccaffery}{Wells
  et~al\mbox{.}}{[n.d.]}]%
        {Wells_chapter17}
\bibfield{author}{\bibinfo{person}{Nancy Wells}, \bibinfo{person}{Chris
  Pasero}, {and} \bibinfo{person}{Margo Mccaffery}.}
  \bibinfo{year}{[n.d.]}\natexlab{}.
\newblock \bibinfo{title}{Chapter 17. Improving the Quality of Care Through
  Pain Assessment and Management}.
\newblock
\newblock


\bibitem[\protect\citeauthoryear{Werner, Lopez-Martinez, Walter, Al-Hamadi,
  Gruss, and Picard}{Werner et~al\mbox{.}}{2019}]%
        {werner2019automatic}
\bibfield{author}{\bibinfo{person}{Philipp Werner}, \bibinfo{person}{Daniel
  Lopez-Martinez}, \bibinfo{person}{Steffen Walter}, \bibinfo{person}{Ayoub
  Al-Hamadi}, \bibinfo{person}{Sascha Gruss}, {and} \bibinfo{person}{Rosalind
  Picard}.} \bibinfo{year}{2019}\natexlab{}.
\newblock \showarticletitle{Automatic Recognition Methods Supporting Pain
  Assessment: A Survey}.
\newblock \bibinfo{journal}{\emph{IEEE Transactions on Affective Computing}}
  (\bibinfo{year}{2019}).
\newblock


\bibitem[\protect\citeauthoryear{Yang, Andrew, Eichner, Sun, Li, Kong, Ramage,
  and Beaufays}{Yang et~al\mbox{.}}{2018}]%
        {ref-google-keyboard}
\bibfield{author}{\bibinfo{person}{Timothy Yang}, \bibinfo{person}{Galen
  Andrew}, \bibinfo{person}{Hubert Eichner}, \bibinfo{person}{Haicheng Sun},
  \bibinfo{person}{Wei Li}, \bibinfo{person}{Nicholas Kong},
  \bibinfo{person}{Daniel Ramage}, {and} \bibinfo{person}{Fran{\c{c}}oise
  Beaufays}.} \bibinfo{year}{2018}\natexlab{}.
\newblock \showarticletitle{Applied Federated Learning: Improving Google
  Keyboard Query Suggestions}.
\newblock \bibinfo{journal}{\emph{CoRR}}  \bibinfo{volume}{abs/1812.02903}
  (\bibinfo{year}{2018}).
\newblock
\showeprint[arxiv]{1812.02903}
\urldef\tempurl%
\url{http://arxiv.org/abs/1812.02903}
\showURL{%
\tempurl}


\end{thebibliography}

\end{document}